\documentclass{ws-rv9x6}
\usepackage{subcaption}   
\usepackage[square]{ws-rv-van}   
\usepackage[cal=boondoxo]{mathalfa}
\usepackage{float}
\usepackage[ruled,vlined]{algorithm2e}

\usepackage{tikz}
\usetikzlibrary{external, shapes}
\definecolor{nicegreen}{HTML}{2CA02C}
\definecolor{darkgreen}{rgb}{0.0,0.5,0.0}

\newcommand{\TC}[1]{#1}

\DeclareMathOperator*{\argmax}{arg\,max}

\newcommand{\w}[0]{\ensuremath{\mathcal{w}}}
\newcommand{\p}[0]{\ensuremath{\mathcal{p}}}
\newcommand{\pa}[0]{\ensuremath{\mathcal{p}_\mathcal{a}}}
\newcommand{\qa}[0]{\ensuremath{\mathcal{q}^\mathcal{m}}}
\renewcommand{\a}[0]{\ensuremath{\mathcal{a}}}
\newcommand{\e}[0]{\ensuremath{\mathcal{e}}}
\newcommand{\D}[0]{\ensuremath{\mathcal{D}}}
\newcommand{\E}[0]{\ensuremath{\mathcal{E}}}
\renewcommand{\S}[0]{\ensuremath{\mathcal{S}}}
\renewcommand{\d}[0]{\ensuremath{\mathcal{d}}}

\newcommand{\x}[0]{\ensuremath{\mathcal{x}}}
\newcommand{\y}[0]{\ensuremath{\mathcal{y}}}

\newcommand{\m}[0]{\ensuremath{\mathcal{m}}}
\renewcommand{\r}[0]{\ensuremath{\mathcal{r}}}
\renewcommand{\t}[0]{\ensuremath{\mathcal{t}}}
\renewcommand{\L}[0]{\ensuremath{\mathcal{L}}}

\makeindex

\begin{document}

\chapter[Bayesian Neural Networks]{Bayesian Neural Networks}\label{ra_ch1}

\author[Charnock, Perreault-Levasseur, Lanusse]{Tom Charnock$^1$, Laurence Perreault-Levasseur$^2$, Fran\c{c}ois Lanusse$^3$}

\address{$^1$Sorbonne Universit\'e, CNRS, UMR 7095, Institut d'Astrophysique de Paris,\\98 bis bd Arago, 75014 Paris, France, \\
tom@charnock.fr}

\address{$^2$Department of Physics, Univesit\'e de Montr\'eal, Montr\'eal, Canada, \\ Mila - Quebec Artificial Intelligence Institute,
Montr\'eal, Canada, and \\
Center for Computational Astrophysics, Flatiron Institute, New York, USA \\
llevasseur@astro.umontreal.ca}

\address{$^3$AIM, CEA, CNRS, Universit\'e Paris-Saclay, Universit\'e Paris Diderot, Sorbonne Paris Cit\'e, F-91191 Gif-sur-Yvette, France \\
francois.lanusse@cea.fr}

\begin{abstract}
In recent times, neural networks have become a powerful tool for the analysis of complex and abstract data models.
However, their introduction intrinsically increases our uncertainty about which features of the analysis are model-related and which are due to the neural network.
This means that predictions by neural networks have biases which cannot be trivially distinguished from being due to the true nature of the creation and observation of data or not.
In order to attempt to address such issues we discuss Bayesian neural networks: neural networks where the uncertainty due to the network can be characterised.
In particular, we outline the Bayesian statistical framework which allows us to categorise uncertainty in terms of the ingrained randomness of observing certain data and the uncertainty from our \emph{lack of knowledge} about how processes that are observed can occur.
In presenting such techniques we show how uncertainties which arise in the predictions made by neural networks can be characterised in principle.
We provide descriptions of the two favoured methods for analysing such uncertainties.
We will also describe how both of these methods have substantial pitfalls when put into practice, highlighting the need for other statistical techniques to truly be able to do inference when using neural networks.
\end{abstract}
\body

\tableofcontents

\section{Introduction}\label{sec:intro}

In recent times we have seen the power and ability that neural networks and deep learning methods can provide for fitting abstractly complex data models.
However, any prediction from a neural network is necessarily and unknowably biased due to factors such as: choices in network architecture; methods for fitting networks; cuts in sets of training data; uncertainty in the distribution of realistic data; and lack of knowledge about the physical processes which generate such data.
In this chapter we elucidate ways in which one can learn how to separate, as much as possible, the sources of error which are due to intrinsic distribution of observed data and those that we have introduced by modelling this distribution both with physical models and by considering neural networks as statistical models.

\TC{
\subsection{The need for statistical modelling}
\label{sec:nsm}

Imagine that we walk into a room and there are ten, six-sided dice whose result we observe. 
The dice are then taken away and we are left to wonder how likely is it that we observed that particular roll. 
Because the dice have been taken away we cannot perform any repeated experiments to make simple estimates of the probability of what we assume is a random process based on counts. 
Instead, we can build a model describing the dice roll and infer the values of the parameters of this model based on how much evidence can be obtained from the observation. 
We could assume that all the dice were equally weighted and there were no external factors to affect the roll and therefore suggest that the result of the dice roll follow a multinomial distribution with equal probability for each of the six possible results from each of the ten dice.
However, what if we had observed nine dice showing one and the other die showing six?
It would be very unlikely to observe such an event within this model, in fact we can calculate the probability of this result in this model to be 0.000017\%.
We could instead decide that each dice is weighted so that there is a 90\% chance that they will land on one and a 10\% chance that they will land on six, in which case the probability of observing this event is much higher at $\sim38\%$.
Or, we could decide that nine of the dice are weighted so that there is a 100\% chance that they will land on one and the tenth die has a 100\% chance that it will land on six, and in which case the observed event is certain.
The problem is that we do not know about the state of the dice or the processes by which different results can be obtained.
Therefore we do not know the values of the parameters in the multinomial model that we use to describe how likely any result is and so there is a source of uncertainty in any prediction we make.
}

\subsection{Aleatoric and epistemic uncertainty}

Uncertainty can be categorised into two classes: aleatoric and epistemic.
These two uncertainties explain, respectively, scatter from what we cannot know and error due to lack of knowledge.
\TC{For example, we do not know how likely it is to have observed nine ones and one six on ten six-sided dice when we do not have access to those dice.}
This is an intrinsic uncertainty due to the random nature of the way the observed event happens and the way we make observations.
As such, we call this uncertainty \emph{aleatoric} since it cannot be reduced through greater understanding.
On the other hand, when we are trying to understand a particular set of observations there are things we do not know but could, in principle, learn about: what are the properties of a physical processes which are necessary to create such data? what types of distribution could describe how likely were we to see such an observation? and how certain are we that such a model is supported by our data?
\TC{For the dice roll example, we do not know if the dice are weighted, or if weighted dice would better fit the observed result, or if there is something that we are not considering, like whether the dice were placed in a particular way rather than thrown and being the result of a random process.}
By addressing the above questions we can narrow down on the possible ways to describe the observation and learn about the state of how the observation came to be through the use of the available data and so any uncertainty due to the lack of knowledge is reducible.
\TC{Knowledge about the result observed on the rolled dice can allow us to narrow down the possible values of the probabilities in the multinomial model that could produce such an observation, therefore reducing our uncertainty.}
We call this reducible uncertainty \emph{epistemic}.

\medskip
\TC{Whilst a simple example, such as the rolling of dice, seems trivial, it describes perfectly any way of learning about our surroundings using the available data.
Every experiment performed exists in a single universe that has undergone epochs of evolution and its constituent particles and forces have interacted to provide us with what we can observe.
There is therefore aleatoric uncertainty due to the fact that we can only observe this one realisation of our universe, and we cannot observe other universes to increase our knowledge about how likely our universe is to be the way it is.
We can, though, make models which describe the constituents of the universe, the way they interact and the evolution to get what we see today.
Although we do not know how likely the observed data is, we can reduce our uncertainty about the possible models, and its parameters values, which are supported by the data.
In fact, even repetitions within a single experiment are taking place at different locations and times in the same single universe--therefore, it is only an assumption of the model for analysing the repeated experiment that any observations are independent results and is not intrinsic to the data that we observe.}

\medskip
The use of neural network for the analysis of data modifies our data model to include any effects that are introduced by the network.
There is, therefore, intrinsic (aleatoric) uncertainty due to the stochastic nature of the data, and epistemic uncertainty now due to both the lack of knowledge about process generating the data as well as the design of the neural network, the way it is trained, the choice of cost function, etc.
\TC{
Neural networks should therefore be seen as an extended, extra-parameterised physical model for the data, whose parameters can be inferred through the support of data.
This means, to be able to use networks to make scientifically relevant predictions, the epistemic uncertainty must be well understood and be able to be characterised.}

\medskip
For the most part, estimates of how well a neural network generalises are obtained using large sets of validation and testing data and relative agreement then suggests that a neural network ``works''.
However, these neural networks do not address the probability that any prediction coincides with the truth.
There is no separation between aleatoric and epistemic uncertainty and no knowledge of how likely (or well) a new example of data is to provide a realistic prediction.
It is \emph{possible}, though, to quantify this epistemic error caused by our lack of knowledge about the properties of a neural network, and characterising this uncertainty can allow us to perform reasoned inference.
In this chapter we will lay down the formalism for Bayesian neural networks: treating neural networks as statistical models whose parameters are attributed probabilities as a degree of belief which can be logically updated under the support from data.
In such a form, neural networks can be used to make statements of inference about how likely we are to believe the outputs of neural networks, reducing the lack of trust that is inherent in the standard deep learning setup.
We will also show some ways of practically implementing this Bayesian formalism and indicating where some of these implementations have been used in astronomy and cosmology.

\section{Bayesian neural networks}\label{sec:bnn}

In this section we will show how one can use a Bayesian statistical framework to assess both aleatoric and epistemic uncertainty in a model which includes neural networks, and describe how epistemic uncertainty can be reduced under the evidence of supporting data using Bayesian inference.

\subsection{Bayesian statistics}

When speaking of \emph{uncertainty}, we are really describing our lack of knowledge about the truth.
This uncertainty is subjective, in that it is not an inherent property of a problem but rather the way we construct the problem.
If we are uncertain about the results of a particular experiment, we do not know exactly what the result of that experiment will be.
The Bayesian  (or subjective) statistical framework is a scientific viewpoint in which we admit that we do not (and are not able to) know the truth about any particular hypothesis.
Our uncertainty, or our degree of belief in the truth, are attributed probabilities, i.e. hypotheses we believe more strongly are described as being more likely.
Of course, in this construction, probabilities can vary from person to person, since different beliefs can be held by different people.
Without any \emph{prior} knowledge, we are free to believe what we will.
However, by using Bayesian inference, we are able to reduce epistemic uncertainty and update our \emph{a priori} knowledge by obtaining evidence, \emph{a posteriori}.
It is important to realise that, whilst our \emph{a priori} beliefs describe the epistemic uncertainty, this quantification can be artificially small without the support of observations.
If our beliefs are not supported by the evidence then the \emph{a posteriori} probability describing the state of our belief after obtaining evidence will become more uncertain, which is a better characterisation of the state of our knowledge.
Under repeated application of new evidence, we can update our beliefs to hone in on the best supported result.

\subsubsection{Statistical models}

A Bayesian statistical framework is a natural setting to build models with which we can infer the most likely distributions and underlying processes that generate some observable events.
Observations can be thought of as existing in measurable space of possible events, $(\S,\E,\mathcal{P})$.
The first element is the sampling space, $\S$, which describes the set of all possible outcomes for a given problem\footnote{For a more in depth discussion of the measure theoretic definition of probability see works such as \citet{athreya2006measure} or other graduate level texts.}.
Each outcome is a random variable, $\d\in\S$, whose value is a single measured observation or result.
An event, $\D\subset\S$, is defined as a subset of all possible outcomes.
The set of all events that can possibly occur is $\E$.
\TC{Referring back to section~\ref{sec:nsm}, we can think of the sampling space, $\S$, as the set of any possible roll of a six-side dice and the value of any roll as an outcome denoted by the value of $\d\in\S$.
An event, $\D$, could then be a collection of different outcomes, such as the nine ones and one six observed on the ten dice described in section~\ref{sec:nsm}.}
A statistical description of data also has a measure on the space of possible events, $\mathcal{P}: \D \in \E \mapsto \mathcal{P}(\D) \in [0,1]$, which is a function that assigns a value between 0 and 1 to every event, $\D\in\E$, describing how likely it is for such an event to occur.
This probability indicates that an event, $\D$, is impossible when $\mathcal{P}(\D)=0$ and is certain when $\mathcal{P}(\D)=1$.
The measure, $\mathcal{P}$, of this measurable space is additive, so that it is certain that any \emph{possible} event \emph{can} occur, $\mathcal{P}(\E)=1$.

\medskip

Whilst we can observe some subset of all possible outcomes from this probability space we do not necessarily know \emph{which} particular outcomes we will observe from the random processes generating any $\mathcal{D}$.
That is, given the value of some observed event $\D$, sampled from $\E$ with a probability $\mathcal{P}$, i.e. from the measurable space $(\S, \E, \mathcal{P})$, we would not know which event would occur from the distribution, $\mathcal{P}$.
\TC{Even if we knew exactly about the dice, how they were weighted, how hard they were thrown, etc. we would still not know exactly which result we would observe if the process had some random aspect.}
The uncertainty due to the statistical nature of the data generation cannot be reduced or learned about and is therefore \emph{aleatoric uncertainty}.

\medskip
\TC{It is the endeavour of science to find models which allow us to describe the things we observe and therefore be able to make predictions using these models.}
In practice, we cannot know the form of $\mathcal{P}$ and as such we attempt to model the probability measure using a statistical model $(\S_\alpha,\E_\alpha,\p)$.
In a Bayesian context, $\S_\alpha$ is another sampling space of possible parameterised distributions with an outcome, $\alpha\in\mathcal{S}_\alpha$, representing all properties of a particular distribution, i.e. functional form, shape, as well as the possible values of some unobservable random variables, $\omega\in\Omega_\alpha$, which generate $\d\in\S$, etc.
Any possible set of $\alpha\in\S_\alpha$ is an event in the space of possible distributions, $\a\in\E_\alpha$, which can model $(\S,\E,\mathcal{P})$.
Effectively, any $\a\in\E_\alpha$ defines a model $((\S,\Omega_\a), (\E,\E_\omega), \pa)$ of $(\S,\E,\mathcal{P})$.
That is, any $\a\in\E_\alpha$ introduces a sampling space of unobservable random variables, $\Omega_\a$, whose values, $\omega\in\Omega_\a$, can generate outcomes, $\d\in\S$.
\TC{$\E_\omega$ then defines the }set of all possible unobservable random variables, $\w\subset\Omega_\alpha$, which can generate events, $\D\in\E$. 
\TC{Considering the dice rolling problem, $\a\in\E_\alpha$ could be the use of a multinomial to model the probability of the distribution of possible results, $\D$, from throwing 10 dice.
In this case, one choice of $\a$ could be that, say, there are six model parameters per die, $\w\in\E_\omega$, which ascribe the probability that each side of each die would land face up.}
Any $\a\in\E_\alpha$ also defines a probability measure, $\pa: (\D,\w) \in (\E,\E_\omega) \mapsto \pa(\D,\w) \in [0,1]$, describing how likely any observable-event-and-unobservable-parameter pairs are\TC{, i.e. how likely any value of the parameters, $\w$, is to give rise to some observation, $\D$, is described by the value of the joint distribution of observables and parameters, $\pa(\D,\w)$. 
}
The possible parameterised statistical model characterises what physical processes generate an observable outcome, our assumption about the possible values of the parameters of those physical processes, and how likely we are to obtain any set of outcomes \emph{and} physical parameters.

\medskip
We assign a probabilistic degree of assumption about the possible models from our \emph{prior} knowledge, $\p : \a\in\E_\alpha\mapsto\p(\a)\in[0, 1]$, that any set of possible distributions, $\a\in\E_\alpha$, encapsulates the underlying probability measure, $\mathcal{P}$, describing the probability of events, $\D$, occurring.
\TC{For example we might believe, thanks to our prior knowledge of the problem, that a model, $\a^*\in\E_\alpha$, describing the probability of outcomes of dice roll as a multinomial distribution with equal parameter values is more likely to be correct than another model, $\a^\dagger\in\E_\alpha$, which uses, say, a Dirichlet distribution.
In this case we would ascribe the probability of $\a^*$ as being more likely than $\a^\dagger$, i.e. $\p(\a^*)>\p(\a^\dagger)$.}
The lack of knowledge about the possible values of $\a$ is the source of \emph{epistemic} uncertainty.
Whilst we will never know the exact distribution of data from $(\S,\E,\mathcal{P})$, we can increase our knowledge about how to model it with $\a\in\E_\alpha$ under the evidence of observed events, $\D\in\E$, thereby reducing the epistemic uncertainty.

\medskip

Since the unobservable parameters, $\w\in\E_\omega$, generate possible sets of observable outcomes, $\D\in\E$, we can write down how likely we are to observe some event, $\D$, \emph{given} that the unobservable parameters, $\w$, have a particular value,
\begin{equation}
	\pa(\D,\w) = \L(\D|\w)p_\a(\w).
\end{equation}
We call $\L : (\D,\w) \in (\E,\E_\omega)\mapsto \L(\D|\w)\in[0, 1]$ the \emph{likelihood} of some values of observables $\D$, given the values of parameters, $\w$, and $p_\a : \w\in\E_\omega\mapsto(\w)\in[0, 1]$ is the \emph{a priori} (or prior) distribution of parameters describing what we assume the values of $\w$ to be based on our current knowledge.
Therefore, some (but not all) of the epistemic uncertainty is encapsulated by $p_\a$.
The prior distribution, $p_\a$, does not, however, describe the form of the parameterised joint distribution, $\pa$, modelling, $(\S,\E,\mathcal{P})$, and so we must also consider how likely is it that we assume our choice of possible distributions, $\p(\a)$, to properly characterise the epistemic uncertainty.

\subsubsection{Bayesian inference}

By observing events, $\mathcal{D}\in\E$, we can update our assumptions about the values of any set of unobservable random variables, $\w\in\E_\omega$, and distributions, $\a\in\E_\alpha$, correctly modelling the probabilistic space, $(\S,\E,\p)$, for some problem.
This is how we can reduce our epistemic uncertainty.
The probability describing our choice of assumptions in the possible values of $\w$ and $\a$ obtained \emph{after} we have observed an event, $\mathcal{D}$, is called the \emph{a posteriori} (or posterior) distribution, $\rho : (\D,\w)\in(\E,\E_\omega)\mapsto\rho(\w|\D)\in[0, 1]$ and can be derived by expanding the joint distribution
\begin{align}
	\pa(\D,\w)\p(\a)&=\L(\D|\w)p_\a(\w)\p(\a)\nonumber\\
    &=\rho(\w|\D)\e(\D)\p(\a),
\end{align}
and equating both sides to get Bayes' theorem
\begin{equation}
  \rho(\w|\D) = \frac{\L(\D|\w)p_\a(\w)}{e(\D)}.\label{eq:bayes}
\end{equation}
This equation tells us that, given a particular parameterised model, $\a$, the probability that some parameters, $\w$, have a particular value when some event, $\D$, is observed is proportional to the likelihood of the observation of such an event given a particular value of the parameters, $\w$, generating the event.
The probability of those parameter values is described by our belief in their value, $p_\a(\w)$.
The \emph{evidence}, $\e:\D\in\E\mapsto\e(\D)\in[0, 1]$, that the parameterised distribution accurately describes the distribution of some event is
\begin{equation}
	\e(\D)=\int_{\E_\omega}d\w\,\L(\D|\w)p_\a(\w).\label{eq:evidence}
\end{equation}
If the probability of $\D$ is small when the likelihood is integrated over all possible sets of parameter values, $\w\in\E_\omega$, both of which are defined by $\a$, then there is little support for that choice of a value of $\a\in\E_\alpha$.
This would suggest that we need to update our assumptions about the parameterised distribution, $\p(\a)$, being able to represent the true model, $(\S,\E,\mathcal{P})$.

\paragraph{Maximum likelihood estimation} In classical statistics, the unobserved random variables, $\w\in\E_\omega$, are considered to be fixed parameters of a particular statistical model, $\a\in\E_\alpha$.
The parameters which best describes some event, $\mathcal{D}$, can be found maximising the likelihood function
\begin{equation}
  \widehat{\w}=\underset{\w\in\E_\omega}{\argmax}\,\L(\D|\w).
\end{equation}
Although this point in parameter space maximises the likelihood and can be found fairly easily by various optimisation schemes, it is completely ignorant about both the shape of the distribution, $\L(\D|\w)$, and how likely we think any particular value of $\w$ (and $\a$) are.
This means that the possible parameters values are degenerated to one point and absolute certainty is ascribed to a choice of model and its parameters.
Furthermore, for skewed distributions, the mode of the likelihood can be far away from the expectation value (or mean) of the distribution and therefore the maximum likelihood estimate might not even be representative.
Any epistemic uncertainty in the model is ignored since we do not consider our belief in $\w$, nevermind how likely $\a$ is.

\paragraph{Maximum \emph{a posteriori} estimation} The simplest form of Bayesian inference is finding the maximum \emph{a posteriori} (MAP) estimate, i.e. the mode of the posterior distribution for a given model, $\a$, as
\begin{align}
  \widehat{\w}&=\underset{\w\in\E_\a}{\argmax}\,\rho(\w|\D)\nonumber\\
  &=\underset{\w\in\E_\a}{\argmax}\,\L(\D|\w)p_\a(\w).
\end{align}
Note that, when we think that any values of the model parameters are equally likely, i.e. the prior distribution, $p_\a(\w)$, is uniform, then $\L(\D|\w)\propto\rho(\w|\D)$ and MAP estimation is equivalent to maximum likelihood estimation.
So, whilst MAP estimation is Bayesian due to the addition of our belief in possible parameter values, $p_\a(\w)$, this form of inference suffers in exactly the same way that maximum likelihood estimation does : the mode of the posterior might also be far from the expectation value and not be representative, and all information about the epistemic uncertainty is underestimated because knowledge about the distribution of parameters is ignored.

\paragraph{Bayesian posterior inference} To effectively characterise the epistemic uncertainty, not only should we consider Bayes' theorem \eqref{eq:bayes}, one should work with the marginal distribution over the prior probability of parameterised models
\begin{align}
    \e(\D)&=\int_{\E_\alpha}d\a\,\e(\D)\p(\a),\nonumber\\
    &=\int_{\E_\alpha}\int_{\E_\omega}d\a d\w\,\L(\D|\w)p_\a(\w)\p(\a).
\end{align}
Practically, the space of possible models, $\E_\alpha$, can be infinitely large, although our belief in possible models, $\p(\a)$, does not have to be.
Still, the integration over all possible models often makes the calculation of $\e(\D)$ effectively intractable.
In practice, we tend to choose a particular model and, in the best case (where we have lots of time and computational power) use empirical Bayes to calculate the mode of the possible marginal distributions
\begin{align}
  \hat{\a}&=\underset{\a\in\E_\alpha}{\argmax}\,\e(\D)\p(\a),\nonumber\\
  &=\underset{\a\in\E_\alpha}{\argmax}\,\int_{\E_\omega}d\w\,\pa(\D|\w)p_\a(\w)\p(\a).\label{eq:EB}
\end{align}
As with the MAP estimate of the parameters, $\hat{\a}$ describes the most likely believed model that supports an event, $\mathcal{D}$.
However, again as with the MAP estimate of the parameters, a model, $\a=\hat{\a}$, might have artificially small epistemic uncertainty due to discarding the rest of the knowledge of the distribution.
To be able to correctly estimate this epistemic uncertainty, one must update, logically, the probability of any possible models and parameters based on the acquisition of knowledge.

\subsection{Neural networks formulated as statistical models}\label{sec:smnn}

We can consider neural networks as part of a statistical model.
In this case, we usually think of an observable outcome as a pair of input and target random variable pairs\footnote{Although we discuss pairs $x$ and $y$ suggesting \emph{inputs} and \emph{targets}, note that this notation is generic. For example, for auto-encoders, we would consider the target to be equivalent to the input, and for generative networks we would consider the input to be some latent variables with which to generate some targets, etc.}, $\d=(x,y)\in\S\,$.
An event is then a subset of pairs $\D=(\x,\y)\in\E$ with probability $\mathcal{P}(\x,\y)$.
We can then use a neural network as a parameterised, non-linear function
\begin{equation}
	\r=\mathcal{f}_{\w,\a}(\x)\label{eq:neural_network}
\end{equation}
where $\r$ are considered the parameters of a distribution which models the likelihood of targets given inputs, $\ell(\y|\x,\w)=\L(\x,\y|\w)/\e(\x)$.
The form of the function, i.e. the architecture, the number, value and distribution of network parameters $\w\in\E_\omega$, initialisation of the network, etc. is described by some hyperparameters, $\a\in\E_\alpha$.
The prescription for this likelihood, $\ell(\y|\x,\w)$, can range from being defined as $\ell(\y|\x,\w)\propto\exp[-\Lambda(\y,\r)]$, where $\Lambda(\y,\r)$ is an unregularised loss function measuring the similarity of the output of a neural network, $\r$, to some target\footnote{For example, a classical mean squared loss corresponds to modelling the negative logarithm of the likelihood as a simple standard unit variance diagonal (multivariate) Gaussian with a mean at the neural network output, $\r$.}, $\y$, to parametric distributions such as a mixture of distributions or neural density estimators.

\medskip

When considering a neural network as an abstract function, it can be possible to obtain virtually any value of $\r$ for a given input $\x$ at any values of the network parameters, $\w$, since the network parameters are often unidentifiable \cite{Muller1998} and the functional form of the possible values of $\r$ is very likely infinite in extent and no statement about convexity can be made.
The reason why we use neural networks is because we can carve out parts of useful parameter space which provide the function which describes how to best fit some known data, $(\x,\y)$, using the likelihood, $\ell(\y|\x,\w)$, as defined by the data itself.
We normally describe this set of known data which ascribes acceptable regions of parameter space where the likelihood makes sense as a \emph{training} set, $(\x,\y)_\textrm{train}\in\E$.
However, evaluating the neural network to get $\r=\mathcal{f}_{\w,\a}(\x)$ and assuming that the output, $\r$, has sensible values to correctly define the form of the likelihood of the sampling distribution of targets will often be misleading\footnote{A sensible likelihood for network targets can be created by making the parameters of the network identifiable. One such method is to use \emph{neural physical engines}\cite{Charnock2019NeuralFunction} , where neural networks are designed using physical motivation for the parameters. However, there is a trade-off with this identifiability which comes at the expense of fitting far less complex functions than are usually considered when using neural networks, but far less data and energy is needed to train such models.\label{fn:npe}}.
This statement is true for \emph{any} value of the network parameters, $\w\in\E_\omega$, since most values of $\w$ do not correspond to neural networks which perform the desired function.

\medskip

Having described neural networks as statistical models we can, further, place them in a Bayesian context by associating a probabilistic quantification of our assumptions, $p_\a(\w)$, to the values of the network parameters, $\w\in\E_\omega$, for a network $\a\in\E_\alpha$, which we believe to be able to represent the the true distribution of observed events, $\mathcal{P}(\x, \y)$, with probability $\p(\a)$.
$\p(\a)$ (and the associated $p_\a(\w)$) represent the epistemic uncertainty due to the neural network, whilst the aleatoric uncertainty arises due to the fact that it is not known exactly which $(\x, \y)$ would arise from the statistical model $(\S,\E,\mathcal{P})$.
We can use Bayesian statistics to update our beliefs and obtain posterior predictive estimates of targets, $\y$, based on this information via the posterior predictive distribution
\begin{equation}
    \p(\y| \x)=\int_{\E_\alpha}\int_{\E_\omega} d\a d\w\,\ell(\y|\x,\w)p_\a(\w)\p(\a).\label{eq:post_pred}
\end{equation}
By integrating over all possible parameters for all possible network choices, we obtain a distribution describing how probable different values of $\y$ are, from our model, which incorporates our lack of knowledge.

\medskip

The region where we assume that the parameters allow the network to perform its intended purpose is described by, $\rho\left(\w|(\x,\y)_\textrm{train}\right)$.
This is our first step in the Bayesian inference.
Bayes' theorem tells us
\begin{equation}
  \rho\left(\w|(\x,\y)_\textrm{train}\right)=\frac{\ell\left(\y_\textrm{train}|\x_\textrm{train},\w\right)p_\a(\w)}{\e\left((\x,\y)_\textrm{train}\right)},\label{eq:post_w}
\end{equation}
so that updating our knowledge of the parameters given the presence of a training set allows us to better characterise the probability of obtaining $\y$ from $\x$ with a particular neural network
\begin{equation}
	\mathcal{p}\left(\y|\x,(\x,\y)_\textrm{train}\right)=\int_{\E_\alpha}\int_{\E_\omega}d\a d\w\,\ell(\y|\x,\w)\rho\left(\w|(\x,\y)_\textrm{train}\right)\p(\a).\label{eq:post_pred_post_w}
\end{equation}
To encapsulate the uncertainty in the network we need to calculate the posterior distribution of network parameters, $\w$, as in~\eqref{eq:post_w}, which we can then use to calculate the distribution of possible $\y$ as described by the predicted $\r$ from the network, as in~\eqref{eq:post_pred_post_w}.
Attention must be paid to the initial choice of $p_\a(\w)$ which still occurs in~\eqref{eq:post_w}\footnote{Historically the choice of prior on the weights has normally been chosen to make the gradients of the likelihood manageable, but this may not be the best justified.
Such a choice in prior could be made more meaningful by designing a model where parameters having meaning (see footnote~\ref{fn:npe}).
Another way to solve this problem is not to consider Bayesian neural networks, but instead transfer the prior distribution of network parameters to the prior distribution of data, $\mathcal{P}(\x,\y)$\citep{Hafner2018} .
Note that, in any case, the prior distribution of data should be considered for a fully Bayesian analysis.}.

\medskip

This description of Bayesian neural networks, therefore, refers solely to networks which are part of a Bayesian model\footnote{There is a common misuse of the term Bayesian neural networks to mean networks which predict posterior distributions, say some variational distribution characterised by a neural density estimator for targets, $\ell(\y|\x,\w)$, but these networks are not providing the true posterior distribution of the target, rather they are simply a fitted distribution approximating (to an unknown degree) the posterior (see section~\ref{sec:classicalnn}).}, i.e. networks where the epistemic uncertainty in the network parameters are characterised by probability distributions, $\rho(\w|\x,\y)$, and thus we are interested in the inference of $\w$.
There are several approaches which are effective for characterising distributions, but each of them have their pros and cons.
In section~\ref{sec:PI}, we present some numerically approximate schemes using the exact distributions and some exact schemes using approximate distributions, these fall under the realms of Monte Carlo methods and variational inference.

\subsubsection{Limitations of the Bayesian neural network formulation}

The goal of a Bayesian neural network is to capture epistemic uncertainties. 
In the absence of any data, the behaviour of the model is only controlled by the prior, and should produce large epistemic uncertainties (high variance of the model outputs) for any given input. 
We then expect that as we update the posterior of network parameters with training data, the epistemic uncertainties should decrease in the vicinity of these training points, as the model is now at least somewhat constrained, but the variance should remain large for Out-Of-Distribution (OOD) regions far from the training set. 
This is the behaviour that one would expect, however, we want to highlight that nothing in the BNN derivation presented in this section necessarily implies this behaviour in practice.

\medskip

As in any Bayesian model, the behaviour of a Bayesian neural network when data is not constraining is tightly coupled to the choice of prior. 
However the priors typically used in BNNs are chosen based on practicality and empirical observation rather than principled considerations on the functional space spanned by the neural network.
There is indeed little guarantee that a Gaussian prior on the weights of a deep dense neural network implies any meaningful uncertainties away from the training distribution. 
In fact, it is easily shown\citep{Hafner2018} that putting priors on weights can fail at properly capturing epistemic uncertainties, even on very simple examples.

\subsubsection{Relation to classical neural networks}\label{sec:classicalnn}

Since neural networks are, in general, able to fit arbitrarily complex models when large enough, we might be able to justify a relatively narrow prior on the hyperparameters, $\p(\a)\approx\delta(\a-\hat{\a})$, meaning that we think that an arbitrarily complex network can encapsulate the statistical model $(\S,\E,\mathcal{P})$\footnote{In assuming $\p(\a)\approx\delta(\a-\hat{\a})$ we are of course neglecting a source of epistemic uncertainty. One possible way that allows us to attempt to characterise the distribution of some subset of $\a$ is the use of Bayesian model averaging or ensemble methods\citep{NIPS2017_7219} . This could be used to sample randomly, for example, from the initialisation values of network parameters or the order with which minibatches of data are shuffled, all of which can affect the preferred region of network parameter space which fits the intended function.}.
Marginalising over the possible hyperparameters gives us
\begin{align}
  \p(\x,\y,\w)&=\int_{\E_\alpha}d\a\,\pa(\x,\y,\w)\p(\a)\nonumber\\
  &=\int_{\E_\alpha}d\a\,\pa(\x,\y,\w)\delta(\a-\hat{\a})\nonumber\\
  &=\p_{\hat{\a}}(\x,\y,\w).
\end{align}
This describes the probability of possible input-target pairs and network parameters for any given choice of hyperparameters, from which we can write
\begin{equation}
    \p\left(\y|\x, (\x,\y)_\textrm{train}\right)=\int_{\E_\omega}d\w\,\hat{\ell}(\y|\x, \w)\hat{\rho}\left(\w|(\x,\y)_\textrm{train}\right),\label{eq:post_pred_w}
\end{equation}
where $\hat{\ell} = \ell|_{\a=\hat{\a}}$ and $\hat{\rho}=\rho|_{\a=\hat{a}}$.

\medskip

In a non-Bayesian context, having restricted the possible forms of neural networks via fixing $\a=\hat{\a}$, it is common to find the mode of the distribution of neural network parameters, $\w$, by maximising the likelihood\footnote{As described earlier, an unregularised loss function can be used to evaluate the negative logarithm of likelihood. A regularisation term on the network parameters can be added describing our belief in how the weights should behave. In this case the regularised loss is proportional to the negative logarithm of the posterior distribution and maximising the regularised loss is equivalent to MAP estimation.} of observing some training set $(\x,\y)_\textrm{train}\in\E$ when given those parameters
\begin{align}
  \widehat{\w}&=\underset{\w\in\E_\omega}{\argmax}\,\hat{\ell}(\y_\textrm{train}|\x_\textrm{train},\w).
\end{align}

\medskip

Once an estimate for the network parameters is made, the posterior distribution of parameter values, $\hat{\rho}\left(\w|(\x,\y)_\textrm{train}\right)$, is usually degenerated to a delta function at the maximum likelihood estimate of the network parameters, $\hat{\rho}\left(\w|(\x,\y)_\textrm{train}\right)\Rightarrow\delta(\w-\widehat{\w})$.
The prediction of a target, $\y$, from an input, $\x$, then occurs with a probability equal to the likelihood evaluated at the maximum likelihood estimate of the value of the network parameters
\begin{align}
  \p\left(\y|\x, (\x,\y)_\textrm{train}\right)&=\int_{\E_\omega}d\w\,\hat{\ell}(\y|\x,\w)\hat{\rho}\left(\w|(\x,\y)_\textrm{train}\right)\nonumber\\
  &=\int_{\E_\omega}d\w\,\hat{\ell}(\y|\x,\w)\delta(\w-\widehat{\w})\nonumber\\
  &=\hat{\ell}(\y|\x,\widehat{\w}).\label{eq:classical_network}
\end{align}
Once optimised, the form of the distribution chosen to evaluate the training samples, i.e. the loss function, is often ignored and the network output, $\r$, is assumed to coincide with the truth, $\y$.
Note, however, that the result of~\eqref{eq:classical_network} is actually a distribution, characterised by the loss function or a variational distribution, at $\w=\widehat{\w}$, peaked at whatever is dictated by the output of the neural network (and not necessarily the true value of $\y$).
Therefore, even in the classical case, we can make an estimation of how likely targets are by evaluating the loss function for different $\y$ using frameworks such as Markov methods (described in section~\ref{sec:mcmc}) or fitting the variational distribution for $\p(\y|\x,\w)$ (described in section~\ref{sec:VI}).

\medskip

However, this form of Bayesian inference does not characterise the uncertainties due to the neural network.
Using the maximum likelihood of the network parameters (and hyperparameters) as degenerated prior distributions for calculating the posterior predictive distribution, $\p\left(\y|\x, (\x, \y)_\textrm{train}\right)$ completely ignores the epistemic uncertainty introduced by the network by assuming that the likelihood with such parameters exactly describes the distribution of $\y$ given a value of $\x$.
Again, even though the value of $\w=\widehat{\w}$ that maximises the likelihood can be found fairly easily by various optimisation schemes, information about the shape of the likelihood is discarded and therefore may not be supported by the bulk of the probability.
To incorporate our lack of knowledge and build true Bayesian neural networks, we have to revert back to~\eqref{eq:post_pred_post_w}.

\section{Practical implementations}\label{sec:PI}

The methods laid out in this chapter showcase some practical ways for characterising distributions.
These distributions could be, for example, the posterior distribution of network parameters, $\rho(\w|\x,\y)$, necessary for performing inference with Bayesian neural networks, or likewise, the predictive density of targets from inputs, $\p(\y|\x)$, normally considered in model inference or, indeed, any other distribution.
For simplicity we will refer, abstractly, to the target distribution as
\begin{equation}
	\rho(\lambda|\chi)=\frac{\L(\chi|\lambda) p(\lambda)}{\e(\chi)}
\end{equation}
for variables $\lambda\in\E_\Lambda$ and observables $\chi\in\E_X$.

\subsection{Numerically approximate inference:\newline Monte Carlo methods}\label{sec:mcmc}

Monte Carlo methods define a class of solutions to probabilistic problems.
One particularly important method is Markov chain Monte Carlo (MCMC) in which a Markov chain of samples is constructed with such properties that the samples can be attributed as belonging to a target distribution.

\medskip
A Markov chain is a stochastic model of events where each event depends on \emph{only} one previous event.
For example, labelling an event as $\lambda_i\in\E_\Lambda$, the probability of transitioning to another event, $\lambda_{i+1}\in\E_\Lambda$ is given by a transition probability, $\t: \{\lambda_i, \lambda_{i+1}\}\in\E_\Lambda\mapsto \t(\lambda_{i+1}|\lambda_i)\in[0, 1]$, where its value describes how likely $\lambda_{i+1}$ will be transitioned to from $\lambda_i$ .
A chain consists of a set of events, called samples, of the state, $\{\lambda_i|\,i\in[1, n]\}$, in which each consecutive sample is correlated with the next.
Although the transition probability is only conditional on the previous state, the chains are correlated over long distances.
Only states that are physically uncorrelated can be kept as samples from some target distribution, $\rho$.

\medskip
One property that a Markov chain must have to represent a set of samples from a target distribution, is \emph{ergodicity}.
This means that it is possible to move from any possible state to another in some finite number of transitions from one state to the next and that no long term repeating cycles occur in the chain.
The stationary distribution of the chain, in the asymptotic limit of infinite samples, can be denoted $\pi : \lambda\in\E_\Lambda\mapsto\pi(\lambda)\in[0, 1]$.
Since an infinite number of samples are needed to prove the stationary condition, MCMC techniques can only be considered numerical approximations to the target distribution.
It should be noted that the initial steps in any Markov chain tend to be out of equilibrium and as such those samples can be out of distribution.
All the samples until the stationary distribution is reached are considered \emph{burn-in} samples and need to be discarded in order not to skew the approximated target distribution.

\subsubsection{Metropolis-Hastings algorithm}\label{sec:MH}

The Metropolis-Hastings algorithm is a methods which allows states to be generated from a target distribution, $\rho$, by defining transition probabilities between states such that the distribution of samples, $\pi$, in a Markov chain is stationary and ergodic.
This can be ensured easily by invoking \emph{detailed balance}, i.e. making the transition probability from state $\lambda_i$ to $\lambda_{i+1}$ reversible such that the Markov chain is necessarily in a steady state.
Detailed balance can be written as
\begin{equation}
  \pi(\lambda_i)\t(\lambda_{i+1}|\lambda_i)=\pi(\lambda_{i+1})\t(\lambda_i|\lambda_{i+1}),
\end{equation}
which is the probability of being in state $\lambda_i$ and transitioning to state $\lambda_{i+1}$ is equal to the probability of being in state $\lambda_{i+1}$ and transitioning to state $\lambda_i$.

\medskip

As described in section~\ref{eq:bayes}, it can be effectively impossible to characterise a distribution, $\rho$, since the integral necessary for calculating the marginal for any observed data, $\e(\chi)$, can often be intractable.
This isn't a problem when using the Metropolis-Hastings algorithm, thanks to detailed balance.
First, substituting the target distribution, $\pi(\lambda)\approx\rho(\lambda|\chi)$, into the detailed balance equation and rearranging gives
\begin{align}
\frac{\t(\lambda_{i+1}|\lambda_i)}{\t(\lambda_i|\lambda_{i+1})} & = \frac{\rho(\lambda_{i+1}|\chi)}{\rho(\lambda_i|\chi)}\nonumber\\
&=\frac{\L(\chi|\lambda_{i+1})p(\lambda_{i+1})/\e(\chi)}{\L(\chi|\lambda_i)p(\lambda_i)/\e(\chi)}\nonumber\\
&=\frac{\L(\chi|\lambda_{i+1})p(\lambda_{i+1})}{\L(\chi|\lambda_i)p(\lambda_i)}.
\end{align}
The intractable integral cancels out and as such we can work with the unnormalised posterior,
\begin{align}
	\varrho(\lambda|\chi)&\equiv\L(\chi|\lambda) p(\lambda)\nonumber\\
	&=\rho(\lambda|\chi)\e(\chi)\nonumber\\
	&=\p(\chi,\lambda),
\end{align}
such that
\begin{equation}
	\frac{\t(\lambda_{i+1}|\lambda_i)}{\t(\lambda_i|\lambda_{i+1})}=\frac{\varrho(\lambda_{i+1}|\chi)}{\varrho(\lambda_i|\chi)}.
\end{equation}
The Metropolis-Hastings algorithm involves breaking the transition probability into two steps, $\t(\lambda_{i+1}|\lambda_i)=a(\lambda_{i+1}, \lambda_i)s(\lambda_{i+1}|\lambda_i)$, with a conditional distribution, $s:(\lambda_{i+1}, \lambda_i)\in\E_\Lambda\mapsto s(\lambda_{i+1}|\lambda_i)\in[0, 1]$, proposing a new sample and a probability, $a(\lambda_{i+1}, \lambda_i)$, describing whether the new sample is accepted as a valid proposal or not.
Substituting these into the detailed balance equations gives
\begin{equation}
  \frac{a(\lambda_{i+1},\lambda_i)}{a(\lambda_i,\lambda_{i+1})}=\frac{\varrho(\lambda_{i+1}|\chi)s(\lambda_i|\lambda_{i+1})}{\varrho(\lambda_i|\chi)s(\lambda_{i+1}|\lambda_i)}.
\end{equation}
A reversible acceptance probability can then be identified as
\begin{equation}
  a(\lambda_{i+1},\lambda_i)=\min\left[1,\frac{\varrho(\lambda_{i+1}|\chi)s(\lambda_i|\lambda_{i+1})}{\varrho(\lambda_i|\chi)s(\lambda_{i+1}|\lambda_i)}\right],
\end{equation}
such that either $a(\lambda_{i+1},\lambda_i)=1$ or $a(\lambda_i,\lambda_{i+1})=1$\footnote{Whilst a reversible Markov chain enforces stationarity, it also leads to a probability of rejecting samples, which can be inefficient.
Although we will not go into detail here, it is also possible to construct a continuous, directional Markov process which is still ergodic.
In this case every sample from the state will be accepted making the algorithm more efficient for collecting samples - although the computation could be more costly.
One example of such a method is the Bouncy Particle Sampler\cite{PhysRevE.85.026703, Bouchard2015} in which samples are obtained from the target distribution by picking a random direction in parameter space and sampling along a piecewise-linear trajectory until the value of target distribution at that state is less than or equal to the value of the target distribution at the initial state.
At this point there is a Poissonian probability of the trajectory bouncing back along another randomised trajectory, drawing samples along the way.
Such methods are state-of-the-art but mostly untested in the literature on sampling neural networks.}.

\medskip

The algorithm itself has two free choices, the first is the number of iterations needed to overcome the correlation of states in the chain and properly approximate the target distribution, but in principle it should approach infinity.
The second is the choice of the proposal distribution, $s$.
It is often chosen to be a multivariate Gaussian whose covariance can be optimised during burn-in to properly represent useful step sizes in the direction of each element of a state.
This ensures that the Markov chain is a random walk.
A poor choice of proposal distribution can cause extremely inefficient sampling and as such it should be chosen carefully.

\medskip

Whilst, in principle, Metropolis-Hastings MCMC will work in high dimensions, the rejection rate can be high and the correlation length very long.
Above a handful of parameters the computational time of Metropolis-Hastings becomes a limitation, meaning that it is not efficient for sampling high dimensional distributions such as the posterior distribution of neural network parameters.

\subsubsection{Hamiltonian Monte Carlo}

One way of dealing with the large correlation between samples, high rejection rate and small step sizes which occur in Metropolis-Hastings is to introduce a new sampling proposal procedure based on a Gibbs sampling step and a Metropolis-Hastings acceptance step.
In Hamiltonian Monte Carlo (HMC), we introduce an arbitrary \emph{momentum} vector, $\nu$, with as many elements as $\lambda$ has.
We describe the Markov process as a classical mechanical system with a total energy (Hamiltonian)
\begin{align}
  \mathcal{H}(\lambda,\nu) &= \mathcal{K}(\nu)+\mathcal{V}(\lambda)\nonumber\\
  &=\frac{1}{2}\nu^T{\bf M}^{-1}\nu-\log \varrho(\lambda|\chi).
\end{align}
$\mathcal{K}(\nu)$ is a \emph{kinetic energy} with a ``mass'' matrix, ${\bf M}$, describing the strength of correlation between parameters.
$\mathcal{V}(\lambda)$ is a \emph{potential energy} equal to the negative logarithm of the target distribution.
A state, $\mathcal{z}=(\lambda, \nu)$, in the stationary distribution of the Markov chain, $\pi(\lambda,\nu)$, is a sample from the distribution $\p(\lambda,\nu|\chi)=\exp[-\mathcal{H}(\lambda,\nu)]$, found by solving the ordinary differential equation (ODE) derived from Hamiltonian dynamics
\begin{align}
  \dot{\lambda} & = {\bf M}^{-1}\nu\label{eq:HMCw}\\
  \dot{\nu} & = -\nabla\mathcal{V}(\lambda),\label{eq:HMCp}
\end{align}
where the dots are derivatives with respect to some time-like variable, which is introduced to define the dynamical system.
The stationary distribution, $\pi(\lambda,\nu)\approx\mathcal{H}(\lambda,\nu)$, of the Markov chain is separable, $\exp[-\mathcal{H}(\lambda,\nu)]=\exp[-\mathcal{K}(\nu)]\exp[-\mathcal{V}(\lambda)]$, and so $\p(\lambda,\nu|\chi)\propto\rho(\lambda|\chi)\p(\nu)$.
This means that a Gibbs sample of the $i^\textrm{th}$ momentum can be drawn, $\nu_i\sim\p=\mathbb{N}({\bf 0}, {\bf M})$, and by evolving the state $\mathcal{z}_i=(\lambda_i,\nu_i)$ using Hamilton's equations, a proposed sample obtained, $\mathcal{z}_{i+1}=(\lambda_{i+1}, \nu_{i+1})\sim\p_\chi$, i.e. $\lambda_{i+1}$ and $\nu_{i+1}$ are drawn with probability $\p(\lambda_{i+1},\nu_{i+1}|\chi)$.
The acceptance condition for the detailed balance is obtained by computing the difference in energies between the $i^\textrm{th}$ state and the proposed, $(i+1)^\textrm{th}$, state
\begin{equation}
  a(\mathcal{z}_{i+1},\mathcal{z}_i) = \min\left[1, \exp(\Delta\mathcal{H})\right],
\end{equation}
where any loss in total energy $\Delta\mathcal{H}=\mathcal{H}(\lambda_{i+1}, \nu_{i+1})-\mathcal{H}(\lambda_i, \nu_i)$ arises from the discretisation of solving Hamilton's equations.
If the equations were solved exactly (the Hamiltonian is conserved), then every single proposal would be accepted.
It is typical to use $\epsilon$-discretisation (the leapfrog method, see algorithm~\ref{al:leapfrog}) to solve the ODE over a number of steps, $L$, where $\epsilon$ describes the step size of the integrator.
Smaller step sizes result in higher acceptance rate at the expense of longer computational times of the integrator, whilst larger step sizes result in shorter integration times, but lower acceptance.
It is possible to allow for self adaptation of $\epsilon$ using properties of the chain, such as the average acceptance as a function of iteration, and a target acceptance rate, $\delta\in[0, 1]$.
It has been shown that, for HMC, the optimal acceptance rate is $\delta\approx0.65$ and so we can adapt $\epsilon$ to be of this order\cite{alex2010optimal}.
Care has to be taken though, since the initial samples in the Markov chain will be out of equilibrium and so adapting $\epsilon$ in the early iterations can still lead to poor step size later on, and so this adaptation should only be attempted after the burn-in phase.
A priori, it is not known how many steps to take in the integrator and so multiple examples of the HMC may need to be run to tune the value of $L$, which can be very expensive\footnote{Recent work has been done using neural networks to approximate the gradient of target distribution, $\nabla\mathcal{V}(\lambda)$\cite{PhysRevD.100.104023}.
Whilst this could lead to errors if trusted for the whole process, the neural gradients are only used in the leapfrog steps to propose new targets, at which point the true target distribution can be evaluated.
In this case, a poorly trained estimator of the gradient of the target distribution proposes poor states, and as such the acceptance rate drops, but the samples obtained are still evaluated from the actual target distribution and therefore it is unbiased by the neural network.
Furthermore, any rejected states could be rerun numerically (rather than being estimated) and added to the training set to further fit the estimator, potentially providing exponential speed up as samples are drawn.
Note, the gradient of the target distribution could be fit using efficient methods described in section~\ref{sec:VI}.}.

\begin{algorithm}
\SetAlgoLined
\SetKwFunction{Calls}{Calls}
{\bf Input:}~Initial state, $\mathcal{z}=(\lambda,\nu)$; number of steps, $L$; step size, $\epsilon$; mass matrix, ${\bf M}$\;
{\bf Output:}~Proposed state, $\mathcal{z}=(\lambda,\nu)$\;
{\bf Calls:}~Gradient of target distribution, $\nabla\mathcal{V}(\lambda)$\;
\BlankLine
 $\nu\leftarrow\nu-\epsilon\nabla\mathcal{V}(\lambda)/2$\;
 \For{$i\leftarrow1$ \KwTo $L$}{
 	$\lambda\leftarrow\lambda+\epsilon{\bf M}^{-1}\nu$\;
 	\If {$i\ne L$}{
 	  $\nu\leftarrow\nu-\epsilon\nabla\mathcal{V}(\lambda)$\;
 	}
 }
 $\nu\leftarrow\nu-\epsilon\nabla\mathcal{V}(\lambda)/2$\;
 \caption{Leapfrog algorithm ($\epsilon$-discretisation)}
\label{al:leapfrog}
\end{algorithm}

\paragraph{No U-turn sampler\cite{Hoffman2014}} A proposed extension to HMC to deal with the unknown number of steps in the integrator is the No U-turn sampler (NUTs).
Here, the idea is to find a condition which describes whether or not running more steps in the integrator would carry on increasing the distance between the initial sample and a proposed one.
A simple choice of criterion is the derivative with respect to Hamiltonian time of the half squared distance between the current proposed and initial states
\begin{align}
	\mathcal{s}&=\frac{d}{dt}\frac{(\lambda_{i+1}-\lambda_i)\cdot(\lambda_{i+1}-\lambda_i)}{2}\nonumber\\
	&=(\lambda_{i+1}-\lambda_i)\cdot\nu.
\end{align}
If $\mathcal{s}=0$ then this indicates that the dynamical system is starting to turn back on itself, i.e. making a U-turn, and further proposals can be closer to the initial state.
In practice, a balanced binary tree of possible samples is created by running the leapfrog integrator either forwards or backwards for a doubling number of steps (1, 2, 4, 8, ...) where each of these steps is a leaf of the tree, $\mathcal{F}=\{(\lambda^{L\pm}, \nu^{L\pm})|L\in[1, 2,  4, ...]\}$.
When the furthest distance in the trajectory, $\lambda^{\max L+}-\lambda^{\max L-}$, starts to decrease then the computation can be stopped and we can sample from $\mathcal{F}$ via a detailed-balance preserving method.
Such an algorithm can greatly reduce the cost of tuning the number of steps in the integrator, $L$, in the HMC and is therefore highly beneficial when attempting to characterise a target distribution.

\medskip

Thanks to the high acceptance rate and the ability to take large steps to efficiently obtain samples, HMC a is good proposition for numerically approximating the distributions such as the posterior distribution of neural network parameters.
One severe limitation, though, is the choice of the mass matrix, ${\bf M}$.
The mass matrix must be properly defined since it defines the direction and size of steps and correlations between parameters.
It is not easy to choose its value \emph{a priori} and a poor choice can lead to very inefficient sampling.
We present below two methods which deal with the mass matrix\footnote{Note we are not going to discuss relativistic HMC \cite{Lu2017} , where the kinetic energy is replaced with its relativistic form $\mathcal{K}(\nu_j)=\sum_{i=1}^{\textrm{dim }\lambda}m_ic_i^2\sqrt{(\nu_j/m_ic)^2+1}$. Whilst this method is valid for preventing the run-away of particles on very glassy target distributions thanks to an upper bound on the distance able to be travelled per iteration, $m$ and $c$ are (in practice) needed for every momenta in the dimension of $\lambda$. This makes it as difficult a problem as \emph{a priori} knowing the mass matrix, ${\bf M}$, in the classical case.}.

\paragraph{Quasi-Newtonian HMC\cite{Fu2016}} With quasi-Newtonian HMC (QNHMC) we make use of the second order geometric information of the target distribution as well as the gradient.
The QNHMC modifies Hamilton's equations to
\begin{align}
  \dot{\lambda} &={\bf B}{\bf M}^{-1} \nu \\
  \dot{\nu} &=-{\bf B}\nabla\mathcal{V}(\lambda)
\end{align}
where ${\bf B}$ is an approximation to the inverse Hessian derived using quasi-Newton methods, for more details see \citet{Fu2016}.
Obtaining this approximation of the Hessian is extremely efficient because all the necessary components are calculated when solving Hamilton's equations using leapfrog methods as in algorithm~\ref{al:leapfrog}.
Note that the approximate inverse Hessian varies with proposal, but is kept constant whilst solving Hamilton's equations.
The inverse Hessian effectively rescales the momenta and parameters such that each dimension has a similar scale and thus the movement around the target distribution is more efficient with less correlated proposals.
It is easiest to begin with an initial inverse Hessian, ${\bf B}_0=\mathbb{I}$, and allow the adaptation of the Hessian to the geometry of the space.
Note that the mass matrix, ${\bf M}$, is still present to set the dynamical time-like scales of Hamilton's equations along each direction, but the rescaling of the momenta via ${\bf B}$ allows us to be fairly ambiguous about its value.
The optimal mass matrix for sampling is equal to the covariance of the target distribution, but in practice, a diagonal mass matrix with approximately correct variance values for the distribution works well.

\paragraph{Example: Inference of the halo mass distribution function} 
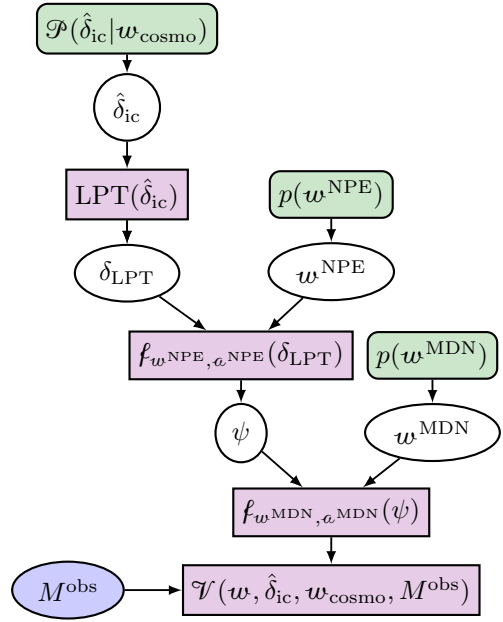
\begin{figure}[!tbh]
        \begin{center}
                \begin{tikzpicture}
    	\pgfdeclarelayer{background}
    	\pgfdeclarelayer{foreground}
    	\pgfsetlayers{background,main,foreground}
    
        \tikzstyle{probability}=[draw, thick, text centered, rounded corners, minimum height=1em, minimum width=1em, fill=darkgreen!20]
    	\tikzstyle{deterministic}=[draw, thick, text centered, minimum height=1.8em, minimum width=1.8em, fill=violet!20]
    	\tikzstyle{variabl}=[ellipse, draw, thick, text centered, minimum height=1em, minimum width=1em]
    	\tikzstyle{data}=[ellipse, draw, thick, text centered, minimum height=1em, minimum width=1em, fill=blue!20]
    
    	\def\blockdist{0.7}
    
        \node (ic) [probability]
        {$\mathcal{P}(\hat{\delta}_\text{ic}|\mathcal{w}_\textrm{cosmo})$};
        \path (ic.south)+(0,-\blockdist) node (deltaic) [variabl]
        {$\hat{\delta}_\text{ic}$};
        \path (deltaic.south)+(0,-\blockdist) node (deltalpt) [deterministic]
        {$\textrm{LPT}(\hat{\delta}_\text{ic})$};
        \path (deltalpt.west)+(3.5,0) node (PNPE) [probability] {$p(\mathcal{w}^\textrm{NPE})$};
        \path (PNPE.south)+(0,-\blockdist) node (thetaNPE) [variabl] {$\mathcal{w}^\textrm{NPE}$};
        \path (deltalpt.south)+(0,-\blockdist) node (deltalpt_field) [variabl]
        {$\delta_\textrm{LPT}$};
        \path (deltalpt_field.south)+(1.5,-\blockdist) node (NPE) [deterministic]
        {$\mathcal{f}_{\mathcal{w}^\textrm{NPE},\mathcal{a}^\textrm{NPE}}(\delta_\textrm{LPT})$};
        \path (NPE.south)+(0,-\blockdist) node (psi) [variabl]
        {$\psi$};
        \path (NPE.west)+(4,0) node (PMDN) [probability] {$p(\mathcal{w}^\textrm{MDN})$};
        \path (PMDN.south)+(0,-\blockdist) node (thetaMDN) [variabl] {$\mathcal{w}^\textrm{MDN}$};
        \path (thetaNPE.south)+(0,-\blockdist-\blockdist-\blockdist-\blockdist) node (MDN) [deterministic] {$\mathcal{f}_{\mathcal{w}^\textrm{MDN},\mathcal{a}^\textrm{MDN}}(\psi)$};
        \path (MDN.south)+(0,-\blockdist) node (likelihood) [deterministic] {$\mathcal{V}(\mathcal{w}, \hat{\delta}_\textrm{ic}, \mathcal{w}_\textrm{cosmo}, M^\textrm{obs})$};
        \path (likelihood.west)+(-1.5,0) node (M) [data] {$M^\textrm{obs}$};
    
    	\path [draw, line width=0.7pt, arrows={-latex}] (ic) -- (deltaic);
    	\path [draw, line width=0.7pt, arrows={-latex}] (deltaic) -- (deltalpt);
    	\path [draw, line width=0.7pt, arrows={-latex}] (deltalpt) -- (deltalpt_field);
    	\path [draw, line width=0.7pt, arrows={-latex}] (deltalpt_field) -- (NPE);
    	\path [draw, line width=0.7pt, arrows={-latex}] (deltalpt) -- (deltalpt_field);
    	\path [draw, line width=0.7pt, arrows={-latex}] (PNPE) -- (thetaNPE);
    	\path [draw, line width=0.7pt, arrows={-latex}] (thetaNPE) -- (NPE);
    	\path [draw, line width=0.7pt, arrows={-latex}] (NPE) -- (psi);
    	\path [draw, line width=0.7pt, arrows={-latex}] (PMDN) -- (thetaMDN);
    	\path [draw, line width=0.7pt, arrows={-latex}] (psi) -- (MDN);
    	\path [draw, line width=0.7pt, arrows={-latex}] (thetaMDN) -- (MDN);
    	\path [draw, line width=0.7pt, arrows={-latex}] (MDN) -- (likelihood);
    	\path [draw, line width=0.7pt, arrows={-latex}] (M) -- (likelihood);

    \end{tikzpicture}
  
        \end{center}
        \caption{Schematic of the BORG algorithm with the neural bias model.
        Initial conditions for the dark matter density field in Fourier space, $\hat{\delta}_\textrm{ic}$, are drawn from a prior given a cosmology $\w_\textrm{cosmo}$, $\mathcal{P}(\hat{\delta}_\textrm{ic}| \w_\textrm{cosmo})$.
        These are then evolved forward using a deterministic prescription, in this example using Lagrangian perturbation theory (LPT).
        The evolved field, $\delta_\textrm{LPT}$, is then transformed further using a neural physical engine, $\mathcal{f}_{\w^\textrm{NPE},\a^\textrm{NPE}}$, whose form is described by $\a^\textrm{NPE}$ and which requires parameters $\w^\textrm{NPE}$ which are drawn from a prior $p(\w^\textrm{NPE})$.
        This provides a field $\psi$ from which the halo mass distribution function can be described using a mixture density network, whose hyperparemeters are $\a^\textrm{MDN}$, with parameters $\w^\textrm{MDN}$ drawn from a prior $p(\w^\textrm{MDN})$.
        This halo mass distribution function can be evaluated at given halo masses to be compared to the masses of haloes $M^\textrm{obs}$ from an observed halo catalogue via a Poissonian likelihood, $\mathcal{V}(\w, \hat{\delta}_{ic}, \w_\textrm{cosmo}, M^\textrm{obs})$.
        The initial phases of the dark matter distribution, $\hat{\delta}_\textrm{ic}$, are sampled using HMC and the parameters of the neural bias model made up of the neural physical engine and the mixture of distributions, $\w=(\w^\textrm{NPE},\w^\textrm{MDN})$, are sampled using QNHMC. \emph{Figure adapted from Charnock et al. (2020)\cite{Charnock2019NeuralFunction}}} 
        \label{fig:mdn_diagram}
    \end{figure}
    
To be able to extract cosmological information from the large scale structure distribution of matter in the universe, such as the mass, location and clustering of galaxies, obtained by galaxy surveys, we either have to summarise the data into statistical quantities (such as the power spectrum, etc.) or learn about the placement of all the objects in these surveys.
Whilst the first method is (potentially very) lossy, the complexity of the likelihood describing the distribution of structures in the universe generally makes the second technique intractable.
With the goal of maximising the cosmological information extracted from galaxy surveys the Aquila consortium has developed an algorithm for Bayesian origins reconstruction from galaxies (BORG)\cite{Jasche:2013, Jasche:2015, Lavaux:2016} which assumes a Bayesian hierarchical model to relate Gaussian initial conditions of the early universe to the complex distribution of galaxies observed today. 
As part of this model, one needs to relate observed galaxies to the underlying, and otherwise invisible, dark matter field through a so-called \emph{bias} model, which is an effective description for extremely complex astrophysical effects. Finding a flexible enough and yet tractable parameterisation for this model \emph{a priori} is a difficult task.
       
\medskip
Using physical considerations, such as locality and radial symmetry, to reduce the numbers of degrees of freedom, a very simple mixture density network with 17 parameters was proposed to model this bias\cite{Charnock2019NeuralFunction}. 
This network, dubbed a neural physical engine due to its physical inductive biases, is small enough that each parameter is exactly identifiable, so that sensible priors could be defined for those parameters. 
The ability to place these meaningful priors on network parameters is well motivated for this physically motivated problem, but may be more difficult to design for problems without physical intuition.

\begin{figure}
 	\centering
		\includegraphics{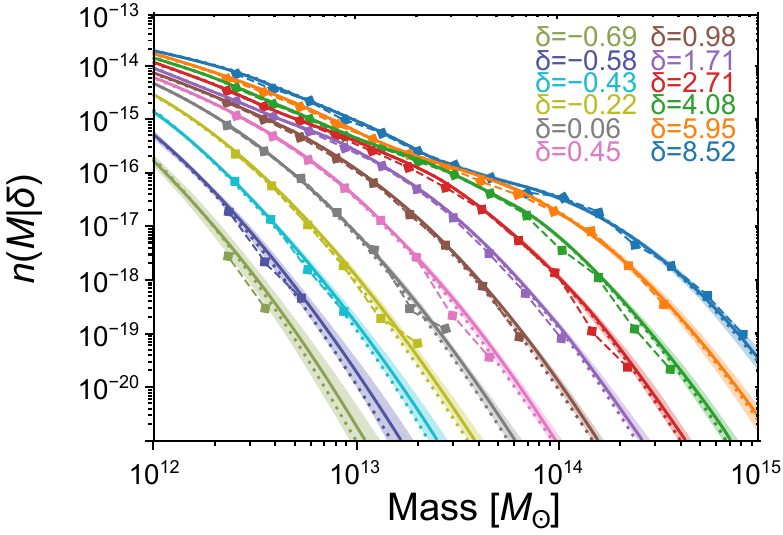}
        \caption{The halo mass distribution function as a function of mass.
        The diamonds connected by a dashed line indicates the number density of haloes from an observed halo catalogue of a given mass, where the different colours represent the value of the density environment for those haloes.
        The lines higher in number density correspond to the more dense regions, i.e. there are more large haloes in denser environments.
        The solid lines show the mean halo number density from samples (taken from the Markov chain) from the neural bias model, with the shaded bands as the 68\% credible intervals of these samples.
        There is a very good agreement between the observed halo number density and that obtained by the neural bias model. \emph{Figure credit: Charnock et al. (2020)\cite{Charnock2019NeuralFunction}}}
        \label{fig:hmdf}
    \end{figure}
\medskip

Sampling from this model could also be integrated within the larger hierarchical model of the BORG framework using QNHMC (see figure~\ref{fig:mdn_diagram} for a description of the BORG+neural bias model algorithm). 
Concretely, BORG was run in two blocks, first using HMC to propose samples of the dark matter density field and then using QNHMC to propose possible neural bias models.
The target distribution was assumed to be a Poissionian sampling of the number density of haloes in any particular environment as described by a mixture density network, $\mathcal{f}_{\w^\textrm{MDN},\a^\textrm{MDN}}(\psi)$, evaluated at possible halo masses, $\mathfrak{m}$, and the summarised environmental properties, $\psi=\mathcal{f}_{\w^\textrm{NPE},\a^\textrm{NPE}}(\delta_\textrm{LPT})$, given by the neural physical engine. 
The parameters in the target likelihood for the masses of haloes in a halo catalogue can be explicitly written in terms of the output of the mixture density network, i.e. $\alpha_{\iota,i_h}\left(\psi(\delta_{\textrm{LPT},i_h},\w^\textrm{NPE}), \w^\textrm{MDN}\right)$, $\mu_{\iota,i_h}\left(\\psi(\delta_{\textrm{LPT},i_h},\w^\textrm{NPE}), \w^\textrm{MDN}\right)$ and $\sigma_{\iota,i_h}\left(\psi(\delta_{\textrm{LPT},i_h},\w^\textrm{NPE}), \w^\textrm{MDN}\right)$ where $i$ labels the number of voxels in the simulator, $h$ labels the number of halos in the catalogue and $\iota$ labels the number of Gaussians in the mixture.
The Poissonian likelihood is then written as
\begin{align}
        \mathcal{V}(\w)  =&\hskip-1.2em \sum_{h\in\textrm{catalogue}}\hskip-1em\log\left[\sum_\iota^N\frac{\alpha_{\iota,i_h}}{\sqrt{2\pi\sigma_{\iota,i_h}^2}}\exp\left[-\frac{\left(\log(\mathfrak{m}_h) - \mu_{\iota,i_h}\right)^2}{2\sigma_{\iota,i_h}^2}\right]\right]\\
         &- V\hskip-0.5em\sum_{i\in\textrm{voxels}}\sum_\iota^N\frac{\alpha_{\iota,i}}{2}\exp\left[\frac{\sigma_{\iota,i}^2}{2}\right]\textrm{erfc}\left[\frac{\log\left(\mathfrak{m}_\tau\right) - \mu_{\iota,i} - \sigma_{\iota,i}^2}{\sqrt{2\sigma_{\iota,i}^2}}\right],\nonumber\label{eq:Ls}
    \end{align}
where $\mathfrak{m}_h$ is the mass of halo $h$, $\mathfrak{m}_\tau$ is a threshold on the minimum mass of halos and $V$ is the volume of a single voxel.
Using this, the exact joint posterior of both density field and network parameters could be inferred under the observation of a mock halo catalogue from the \textsc{velmass} simulations~\cite{2019PhRvD.100d3515K}.
Such inference was an example of zero-shot training since there was no \emph{training} data necessary.
Because Quasi-Netwonian HMC was used then the correlation between all of the network parameters, $\w=(\w^\textrm{NPE}, \w^\textrm{MDN})$, could be assumed to be negligible, i.e. ${\bf M}=\mathbb{I}$.
Whilst this assumption is incorrect, the approximately calculated Hessian, ${\bf B}$, rescaled the parameters such that their correlations were taken into account so that the proposed samples in the HMC could successfully travel along preferred parameter directions.
The methods presented here provide a way to constrain the initial phases of the dark matter distribution, conditional on the observed data, without an exact physical description of how the observed catalogues trace the underlying dark matter distribution today.
This is possible since we can use the neural bias model to map from the dark matter distribution to some unknown function that describes how a catalogue of observed halos traces the underlying dark matter based on some physical principles. 
These physical principles are built directly into the neural physical engine.
Any uncertainty in the form of this description can then be marginalised out since the distribution of the parameters in the neural bias model is available via the samples obtained in the QNHMC.

\paragraph{Riemannian Manifold HMC}\cite{Girolami2009}\label{sec:RMHMC} Whilst we have so far depended on a choice of mass matrix to set the time-like steps in the integrator, it is possible to exploit the geometry of the Hamiltonian to adaptively avoid having to choose.
Samples from the Hamiltonian are effectively points in a Riemannian surface with a metric defined by the Fisher information of the target distribution, $\mathcal{I}(\lambda)=\langle\nabla\varrho(\lambda'|\chi)(\nabla\varrho(\lambda'|\chi))^T\rangle_\lambda$.
Since the Fisher information describes the amount of information a random observable, $\chi$, contains about any model parameters $\lambda$, then the parameter space has larger curvature wherever there is a lot of support from the data.
In essence, this metric is a position-dependent equivalent to the mass matrix which we have so far considered, but since we have to calculate $\nabla\varrho(\lambda|\chi)$ in the integrator anyway, we can actually approximate the Fisher information cheaply.
However, to ensure that the Hamiltonian is still the logarithm of a probability density it must be regularised leaving the Hamiltonian as
\begin{align}
	\mathcal{H}(\lambda,\nu)&=\mathcal{K}(\lambda,\nu)+\mathcal{V}(\lambda)\nonumber\\
	&=\frac{1}{2}\nu^T\mathcal{I}(\lambda)^{-1}\nu-\log\varrho(\lambda|\chi)+\frac{1}{2}\log(2\pi)^{\textrm{dim }\lambda}|\mathcal{I}(\lambda)|.
\end{align}
Here the kinetic term now involves a dependence on the parameters, $\lambda$, and so the Hamiltonian is not separable, i.e. the momenta are drawn from a parameter dependent mass matrix, $\nu\sim\mathbb{N}({\bf 0},\mathcal{I}(\lambda))$.
The equations of motion in this case become
\begin{align}
  \dot{\lambda} & = \mathcal{I}(\lambda)^{-1}\nu\\
  \dot{\nu} & = -\nabla\mathcal{V}(\lambda)+\frac{1}{2}\textrm{Trace}\left[\mathcal{I}(\lambda)^{-1}\nabla\mathcal{I}(\lambda)\right]\nonumber\\
  		&\phantom{somestuff}-\frac{1}{2}\nu^T\mathcal{I}(\lambda)^{-1}\nabla\mathcal{I}(\lambda)\mathcal{I}(\lambda)^{-1}\nu.
\end{align}
With such a change, the scaling of the momenta along each parameter direction becomes automatic, but the reversibility and volume preserving evolution using the leapfrog integrator is broken and so the proposed states do not adhere to detailed balance.
Instead a new symplectic integrator, i.e. an integrator for Hamiltonian systems, is required to make the volume preserving transformation of the momenta (by calculating the Jacobian of the inverse Fisher matrix) such that Hamilton's equations can be solved.
Whilst this adds extra complexity to the equations of motion, it is equivalent to only two additional steps in the integrator since the Fisher information can be approximated cheaply from the calculation of the gradient of the potential energy.
By using the RMHMC, we avoid the need to choose a mass matrix (or approximate the Hessian).
$\epsilon$ can be fixed to some value as the adaptive matrix is able to overcome the step size, and $L$, i.e. the number of steps in the integrator, can be chosen to tune the acceptance rate.

\paragraph{Stochastic gradient HMC}\cite{Chen2014} Whilst we can sample effectively using Hamiltonian Monte Carlo, and its variants shown above, we must also address the question of the size of the data we are interested in.
As so far presented, HMC requires an entire dataset, $\chi\in\E_X$, to evaluate the distribution.
However, in modern times, datasets can be extremely large (big data) and it may not be possible to evaluate it all simultaneously.
Furthermore, we are more and more likely to be in the regime where the data is obtained continually and streamed for inference.
For this reason, most optimisation techniques rely on stochastic estimation techniques over minibatches, $\chi_i$ i.e. the union of all minibatches is the complete set, $\bigcup_i^\textrm{batches}\chi_i=\chi$.
This stochastic sampling of data can be considered as being equivalent to adding a source of random noise, or scatter, around the target distribution.
For clarity, the gradient of the potential energy for a minibatch becomes
\begin{align}
	\nabla\mathcal{V}(\lambda)&=-\log\varrho(\lambda|\chi_i)\nonumber\\
	&\to-\log\varrho(\lambda|\chi)+\gamma
\end{align}
where $\gamma\sim\textrm{Dist}(\lambda)$ is a random variable drawn from the distribution of noise and whose shape is described by some diffusion matrix, $\mathcal{Q}(\lambda)$.
For large minibatch sizes, we can relatively safely assume this distribution is Gaussianly distributed, $\gamma\sim\mathbb{N}({\bf 0},\epsilon\mathcal{Q}(\lambda)/2)$, due to the central limit theorem where the diffusion matrix at any step in the integration can be equated to the variance of the noise, $\epsilon\mathcal{Q}(\lambda)=2\Sigma(\lambda)$.
Note, that we may not, necessarily, be in the regime where we can make this assumption.

\medskip

Making noisy estimates of the target distribution using minibatches brakes the Hamiltonian dynamics of the HMC and as such extremely high rejection rates can occur.
In particular, the additional noise term acts as force which can push the states far from the target distribution.
We can reduce this effect by taking further inspiration from mechanical systems - we can use Langevin dynamics to describe the macroscopic states of a statistical mechanical system with a stochastic noise term describing the expected effect of some ensemble of microscopic states.
In particular, using second-order Langevin equations is equivalent to including a friction term which decreases the energy and, thus, counterbalances the effect of the noise.
Equations~\ref{eq:HMCw} and~\ref{eq:HMCp} therefore get promoted to
\begin{align}
  \dot{\lambda} & = {\bf M}^{-1}\nu\\
  \dot{\nu} & = -\nabla\mathcal{V}(\lambda)-\mathcal{Q}(\lambda){\bf M}^{-1}\nu+\gamma
\end{align}
Solving these equations provides a stationary distribution, $\pi(\lambda,\nu)\approx\mathcal{H}(\lambda,\nu)$, with the distribution of samples, $\lambda\sim\rho_\chi$, i.e. $\lambda$ is drawn with probability $\rho(\lambda|\chi)$.
Of course, this method depends on knowing the distribution of the noise well, but for large minibatch sizes, this approaches Gaussian.
The stochastic gradient HMC, in this case, provides a way to obtain samples from the target distribution even when not using the entire dataset and therefore vastly reducing computational expense and allowing for active collection and inference of data.

\subsection{Variational Inference}\label{sec:VI}

Whilst a target probability distribution can be approximately characterised by obtaining exact samples from the distribution via Monte Carlo methods, it is often a very costly process.
Instead we can use variational inference, where a \emph{variational} distribution, say $\qa(\lambda|\chi)$, is chosen to represent a very close approximation to the target distribution, $\rho(\lambda|\chi)$.
In general, $\qa(\lambda|\chi)$ is a tractable distribution, parameterised by some $\mathcal{m}$, and via the optimisation of these parameters $\qa(\lambda|\chi)$ can hopefully be made close to the target distribution, $\rho(\lambda|\chi)$.
Note, again, that if the target distribution is the posterior predictive distribution of some model, $\p(\y|\x)$, then fitting a variational distribution to this is not a Bayesian procedure in the same way that maximum likelihood estimation is not Bayesian.

\medskip
To describe what is meant by \emph{close} in the context of distributions we often consider the Kullback-Leibler (KL) divergence (or relative entropy).
In a statistical setting, the KL-divergence is a measure of information lost when approximating a distribution $\rho(\lambda|\chi)$ with some other $\qa(\lambda|\chi)$,
\begin{equation}
  \mathbb{KL}(\rho||\qa)=\int_{\E_\Lambda}d\lambda\,\rho(\lambda|\chi)\log\frac{\rho(\lambda|\chi)}{\qa(\lambda|\chi)}.\label{eq:KL}
\end{equation}
When $\mathbb{KL}(\rho||\qa)=0$, there is no information loss and so $\rho(\lambda|\chi)$ and $\qa(\lambda|\chi)$ are equivalent.
Values $\mathbb{KL}(\rho||\qa)>0$ indicate the degree of information lost.
Note that the KL-divergence is not symmetric and as such is not a real distance metric.

\medskip
The form of~\eqref{eq:KL} assumes the integral of $\rho(\lambda|\chi)$ to be tractable.
In fact, in the case that the expectation can be approximated well, we can use the KL-divergence to perform expectation propagation.
However, if we are considering the approximation of the posterior distribution of network parameters, as stated in section~\ref{sec:bnn}, we can expect the integral of $\pa(\w|\x,\y)$ to be intractable meaning that calculating $\mathbb{KL}(\pa||\qa)$ would be necessarily hard.
Instead we can consider the reverse KL-divergence
\begin{equation}
  \mathbb{KL}(\qa||\rho)=\int_{\E_\Lambda}d\lambda\,\qa(\lambda|\chi)\log\frac{\qa(\lambda|\chi)}{\rho(\lambda|\chi)}.
\end{equation}
The choice of $\qa(\lambda|\chi)$ is specified so that expectations are tractable.
However, evaluating $\rho(\lambda|\chi)$ would require calculating the evidence, $\e(\chi)$, which, although constant for different $\lambda$, remains intractable.
For convenience we can consider the unnormalised distribution (as we did for detailed balance in section~\ref{sec:MH})
\begin{align}
	\varrho(\lambda|\chi)&=\L(\chi|\lambda) p(\lambda)\nonumber\\
	&=\rho(\lambda|\chi)\e(\chi)\nonumber\\
	&=\p(\chi,\lambda),
\end{align}
 and calculate a new measure
\begin{equation}
  \textrm{ELBO}(\qa)=-\int_{\E_\Lambda}d\lambda\,\qa(\lambda|\chi)\log\frac{\qa(\lambda|\chi)}{\varrho(\lambda|\chi)}.
\end{equation}
Note that this has the form of minus the reverse KL-divergence, but is not equivalent since $\varrho(\lambda|\chi)$ is not normalised.
By substitution we can see that
\begin{align}
  \textrm{ELBO}(\qa)&=-\int_{\E_\Lambda}d\lambda\,\qa(\lambda|\chi)\log\frac{\qa(\lambda|\chi)}{\rho(\lambda|\chi)\e(\chi)}\nonumber\\
  &=-\int_{\E_\Lambda}d\lambda\,\qa(\lambda|\chi)\log\frac{\qa(\lambda|\chi)}{\rho(\lambda|\chi)}+\log\e(\chi)\nonumber\\
  &=-\mathbb{KL}(\qa||\rho)+\log\e(\chi).\label{eq:ELBO}
\end{align}
Since $\log\e(\chi)$ is constant with respect to the parameters, $\lambda$, maximising $\textrm{ELBO}(\qa)$ will force $\qa(\lambda|\chi)$ close to the target distribution $\rho(\lambda|\chi)$.
The term ELBO comes from the fact that the KL-divergence is non-negative and so $\textrm{ELBO}(\qa)$ defines a lower bound to the evidence, $\e(\chi)$.

\subsubsection{Mean-field variation}

One efficient way of parameterising a distribution for approximating a target, $\rho(\lambda|\chi)$, is to make it factorise along each dimension of the parameters, i.e.
\begin{equation}
  \qa(\lambda|\chi)=\prod_{i=1}^{\textrm{dim}\,\lambda}\qa_i(\lambda_i|\chi).
\end{equation}
In doing such, the ELBO for any individual $\qa_j$ is
\begin{align}
  \textrm{ELBO}(\qa_j) &= \underset{\E_{\Lambda,j}}{\int\cdots\int} d\lambda_j\prod_i\qa_i(\lambda_i|\chi)\times\left[\log\varrho(\lambda|\chi)-\sum_k\log\qa_k(\lambda_k|\chi)\right]\nonumber\\
  &=\int_{\E_{\Lambda,j}}d\lambda_j\,\qa_j(\lambda_j|\chi)\underset{\E_{\Lambda,i\ne j}}{\int\cdots\int}d\lambda_i\prod_{i\ne j}\qa_i(\lambda_i|\chi)\nonumber\\
  &\phantom{biglongline}\times\left[\log\varrho(\lambda|\chi)-\sum_k\log\qa_k(\lambda_k|\chi)\right]\nonumber\\
  &=\int_{\E_{\Lambda,j}}d\lambda_j\,\qa_j(\lambda_j|\chi)\times\left[\underset{i\ne j}{\mathbb{E}}[\log\varrho(\lambda_j|\chi)]-\log\qa_j(\lambda_j|\chi)\right]\nonumber\\
  &\phantom{biglongline}+ \textrm{const}
\end{align}
where the constant is the expectation value of the factorised distributions, $\qa_i(\lambda_i|\chi)$, in the dimensions where $i\ne j$ and is unimportant for the optimisation of the distribution for the $j^{th}$ dimension since it is independent of $\lambda_j$.
$\mathbb{E}_{i\ne j}[\log\varrho(\lambda_j|\chi)]$ is the expectation value of the logarithm of the target distribution for every $\qa_i(\lambda_i|\chi)$ where $i\ne j$, and remains due to its dependence on $\lambda_j$.
The optimal $j^{th}$ distribution is the one that maximises the ELBO which is equivalent to optimising each of the factorised distributions, $\qa_j(\lambda_j|\chi)$, in turn to obtain $\qa_j(\lambda_j|\chi)=\exp\left[\mathbb{E}_{i\ne j}\left[\log\varrho(\lambda_j|\chi)\right]\right]$.
This provides a mean-field approximation of the target distribution.

\subsubsection{Bayes by Backprop}
Bayes by Backprop\citep{2015arXiv150505424B, Dikov2019BayesianArchitectures} (a form of stochastic gradient variational Bayes) provides a method for approximating a target distribution, $\rho(\lambda|\chi)$, using differentiable functions such as neural networks.
The basic premise of Bayes by Backprop relies on a technique known as the \emph{reparameterisation trick}~\cite{2013arXiv1306.0733K, 2013arXiv1312.6114K} .
The reparametrisation trick provides a method of drawing random samples, $\w$, from a Gaussian distribution, whilst allowing the samples to be differentiable with respect to the parameters of the Gaussian distribution (mean and standard deviation, $m=(\mu,\sigma)$). 
This can be achieved by reparameterising $\mathcal{N}(\mu_i, \sigma_i)$ in terms of an independent normally distributed auxiliary variable, $\epsilon\sim \mathcal{N}(0, 1)$, i.e.
\begin{align}
	\w(m_i) &\sim\mathbb{N}(\mu_i, \sigma_i)\nonumber\\
	&=\mu_i + \sigma_i \epsilon_i.
\end{align}
Here, we can view $\w(m_i)$ as the $i^\textrm{th}$ random variable parameter of a neural network with a total of $n_\textrm{w}$ network parameters where $\w(\m)=\{\w(m_i)|\,i\in[1,n_\textrm{w}]\}$.
Any evaluation of the neural network is a sample $\widehat{\lambda}\sim\qa_{\chi,\w}$, i.e. $\widehat{\lambda}$ is drawn with probability $\qa(\lambda|\chi,\w)$.
Maximising the ELBO~\eqref{eq:ELBO}, between $\qa(\lambda|\chi,\w)$ and an unnormalised target distribution, $\varrho(\lambda|\chi)$, can now be done via backpropagation since we can calculate
\begin{equation}
	\partial_\m\textrm{ELBO}(\qa)=\begin{pmatrix}\partial_{\w(\m)}\textrm{ELBO}(\qa)+\partial_{\mu}\textrm{ELBO}(\qa)\\(\epsilon/\sigma)\partial_{\w(\m)}\textrm{ELBO}(\qa)+\partial_{\sigma}\textrm{ELBO}(\qa)\end{pmatrix}
\end{equation}
and update the parameters using
\begin{equation}
	\m\leftarrow\m-\eta\partial_{\m}\textrm{ELBO(\qa)}\label{eq:BbB}
\end{equation}
where $\eta$ is a learning rate.
Note that the $\partial_{\w(\m)}\textrm{ELBO}(\qa)$ terms in~\eqref{eq:BbB} are exactly the same as the gradients normally associated with backpropagation in neural networks.

\medskip

As originally presented, Bayes by Backprop was an attempt to make Bayesian posterior predictions of targets, $\y$, from inputs, $\x$, as in~\eqref{eq:post_pred_post_w}, where $\rho(\w|\x_\textrm{train},\y_\textrm{train})\equiv\prod_{i=1}^{n_\textrm{w}}\mathbb{N}(\mu_i,\sigma_i)$.
Here, the values of all $\mu_i$ and $\sigma_i$ are fit using maximum likelihood estimation (or maximum \emph{a posterior} estimation) given data $\x_\textrm{train}$ and $\y_\textrm{train}$.
As explained in section~\ref{sec:smnn}, the distribution of weights, $p_\a(\w)$, is likely to be extremely non-trivial, since most network parameters are non-identifiable and highly degenerate with other parameters.
Therefore, modelling this distribution as a Gaussian is unlikely to be very accurate.
This can, therefore, incorrectly conflate the epistemic uncertainty for $\y$ from a particular network, $\a$, with parameters, $\w$, and input, $\x$, with the posterior prediction.
In essence, Bayes by Backprop provides a way of sampling from a single choice of an (arbitrarily complex) approximation of a target distribution
much more efficiently than using numerical schemes such as Markov methods, but there is little knowledge in how close this approximation is to the desired target.
By fitting the parameters, $\m$, of the neural distribution rather than characterising their distribution, $\p(\m)$, the characterisation of the epistemic uncertainty is biased.
As such, just using Bayes by Backprop provides a network that is Bayesian in principle, but with a limited choice of prior distribution which may fail to capture our epistemic uncertainty.

\medskip

Whilst Bayes by Backprop allows us to characterise the mean and standard deviation of a Gaussian distribution from which we can draw network parameters, it can be extremely expensive to draw different parameters for each example of the data to perform the optimisation.
Therefore, the data is often split into $n_\textrm{batches}$ minibatches of $n_\textrm{elems}$ elements and a single sample of each parameter drawn for all $n_\textrm{elems}$ elements in each minibatch.
This clearly does not represent the variability of the distribution of parameters well and leads to artificially high variance in the stochastic gradient calculation.
Furthermore, by sharing the same parameter values for all elements in a minibatch, correlations between gradients prevents the high variance from being eliminated.

\paragraph{Example: Classification of photometric light-curves} Because Bayes by Backprop can be comparatively more expensive that other practical techniques introduced below, there are a fairly limited number of examples of applications in the physics literature. One notable example however is the probabilistic classification of SuperNovae lightcurves method \texttt{SuperNNova}\cite{2020MNRAS.491.4277M}. The aim of that study is to analyse time-series measuring the brightness of distant galaxies as a function of time, and detect potential
SuperNovae Type Ia events, of particular interest for cosmology.

\begin{figure}
\centering
\begin{subfigure}[t]{.45\textwidth}
    \centering
    \includegraphics[width=\textwidth]{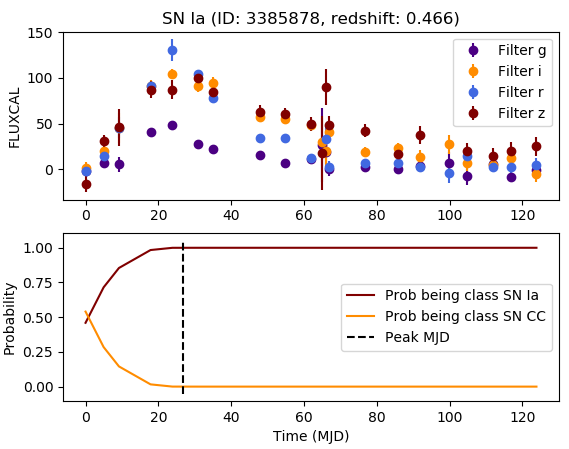}
    \caption{Illustration of the photometric lightcurve classification problem on simulated SuperNova Type Ia event. Top: Observed flux as a function of time, in different bands (broad wavelength filters). Bottom: Classification probability output from a recurrent neural network (RNN) as a function time. While uncertain about the type of the event (Ia or CC) at the beginning of the event, the model starts recognizing a Ia event and classifying it as such as the event unfolds.   \textit{Figure credit: M\"oller \& de Boissiere (2019)\citep{2020MNRAS.491.4277M}}}
    \label{fig:moller2019_lc}
\end{subfigure}
\begin{subfigure}[t]{.1\textwidth}
\end{subfigure}
\begin{subfigure}[t]{.45\textwidth}
    \centering
    \includegraphics[width=\textwidth]{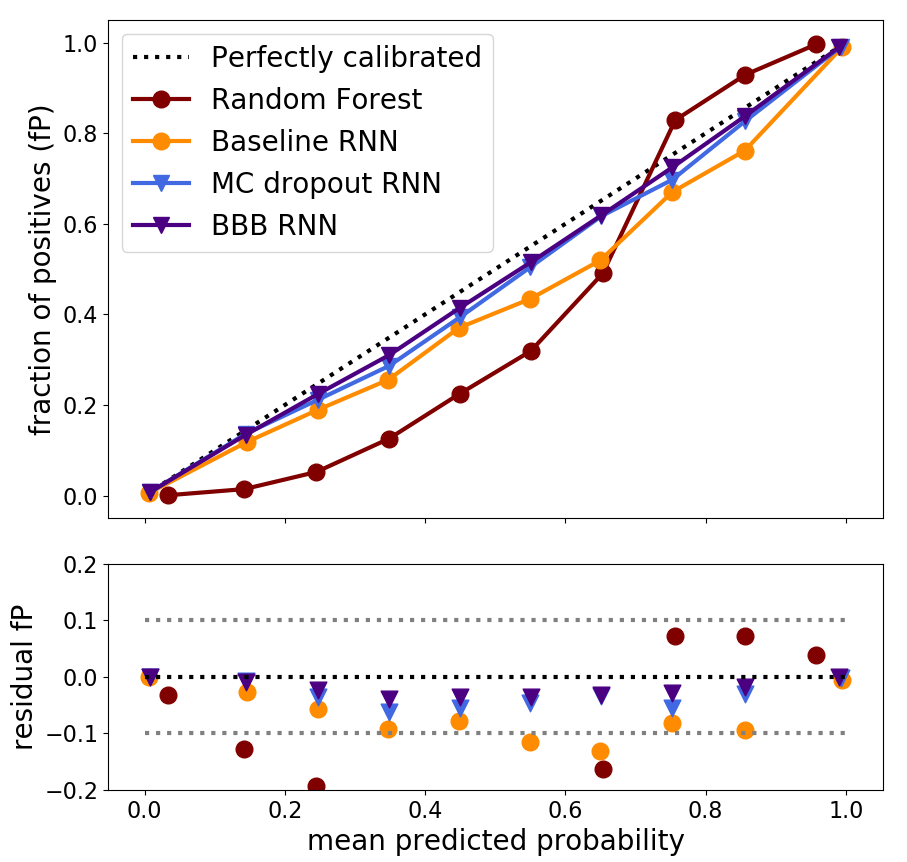}
    \caption{Calibration of predicted class probabilities for the SuperNovae lightcurve classification problem.\newline The BBB RNN (purple triangle) exhibits better calibration than a vanilla RNN (orange dot) of matching architecture but optimised by maximum likelihood. \textit{Figure credit: M\"oller \& de Boissiere (2019)\citep{2020MNRAS.491.4277M}}}
    \label{fig:moller2019_calib}
\end{subfigure}
\label{fig:moller2019}
\end{figure}
    
\medskip 

Figure~\ref{fig:moller2019_lc} illustrates the light curve classification problem on a simulated SuperNova Ia event. In that study, the authors aim to compare the performance of a vanilla recurrent neural network (RNN) classifier to a probabilistic model quantifying some uncertainties. For that purpose the authors introduce a ``\emph{Bayesian}'' recurrent neural network\citep{Fortunato2017} (BRNN) based on a bidirectional LSTM, but using a variational Gaussian distribution for the posterior of network parameters, optimised by back-propagation. Following the approach presented in this section, the loss function for this model becomes
\begin{equation}
	\Lambda(\m) = -\underset{\w\sim\qa}{\mathbb{E}}\left[ \log \ell(\y_\textrm{train} | \x_\textrm{train}, \w) \right] + \mathbb{KL}(\qa || \rho)
\end{equation}
which corresponds to the ELBO introduced in \eqref{eq:ELBO}, and where $\log \ell (\y_\textrm{train} | \x_\textrm{train}, \w)$ is the log-likelihood of a categorical distribution with probabilities predicted by the neural network, and $p_\a(\w)$ is the prior on the BRNN parameters.

\medskip

With this approach, the authors attempt to distinguish between aleatoric uncertainties which are uniquely determined by the categorical probabilities predicted by the model for a given set of network parameters $\w$, and the epistemic uncertainties which are, this case, characterised by the variational approximation to the posterior of network parameters $\qa(\w|(\x,\y)_\textrm{train})\approx\rho(\w|(\x,\y)_\textrm{train})$. As highlighted multiple times before, one should however always be careful in interpreting these probabilities, and in that study the authors empirically check the calibration of the mean posterior probabilities using a reliability diagram\cite{degroot1983} . The reliability diagram shows the fraction of true positives in a binary classification problem as a function of the probabilities predicted by the model. For a perfectly calibrated classifier, only 10\% of objects which received a detection probability of 0.1 are true positives.

\medskip

Figure~\ref{fig:moller2019_calib} shows this calibration diagram for different classifiers, but of particular interest are the curves for the \textit{baseline RNN} and \textit{BBB RNN}, in both cases the actual neural network architecture is identical, but the former is optimised to find the maximum likelihood estimates of the network parameters, while the later is trained by Bayes by Backprop. The \textit{BBB RNN} predicted probabilities are closer to the diagonal representing a more \emph{correct} calibration than the baseline RNN. In this example including a model for the epistemic uncertainties improves the model calibration.

\subsubsection{Local reparameterisation trick}

Although for large numbers of network parameters, $n_\textrm{w}$, characterising the global uncertainty of the parameters using the reparameterisation trick becomes computationally unfeasible for each element of data and for each parameter in the network, a local noise approximation\cite{KingmaVariationalTrick} can be made to transform the perturbation of parameters to a perturbation of activation values
\begin{equation}
	o^l_{jn}\sim\mathbb{N}\left(\sum_{i=1}^{\textrm{dim }l-1}\mu^l_{ji}a_{in}^{l-1},\sum_{i=1}^{\textrm{dim }l-1}\left(\sigma_{ji}^{l}\right)^2\left(a_{in}^{l-1}\right)^2\right),
\end{equation}
where $a_{jn}^l=\mathcal{f}(o_{jn}^l)$ is the activated output (with possibly non-linear activation function $\mathcal{f}$) of the $j^\textrm{th}$ unit of the $l^\textrm{th}$ layer of a neural network with $n_\textrm{l}$ layers, according to the $n^\textrm{th}$ element of the input minibatch.
As with the reparameterisation trick, the sampling of the activation value of any layer can be written as
\begin{equation}
	o^l_{jn}=\sum_{i=1}^{\textrm{dim }l-1}\mu^l_{ji}a_{in}^{l-1}+\epsilon^l_{jn}\sqrt{\left(\sigma_{ji}^{l}\right)^2\left(a_{in}^{l-1}\right)^2}
\end{equation}
where $\epsilon^l_{jn}\sim\mathbb{N}(0, 1)$.
Whilst the parameters $m_{ji}^l=(\mu_{ji}^l,\sigma_{ji}^l)$ of the Gaussian distribution describe the probabilistic model for a network parameter, $\w(m_{ji}^l)$, from unit $i$ of layer $l-1$ to unit $j$ of layer $l$, this model is never sampled, and only the activation values are sampled.
The dimensionality of the probabilistic interpretation of layer outputs, i.e. the number of $\boldsymbol{\epsilon}=\{\epsilon_{in}^l|\,i\in[1,\textrm{dim }l],l\in[1,n_\textrm{l}],n\in[1,n_\textrm{elems}]\}$ needed to be stored for computation of the gradient is much lower than when considering the number of random draws needed for every single network parameter and every element of data in the minibatch.
Furthermore, the variance of the gradient is much less when using the local reparameterisation trick than when assuming a single random draw for each parameter being the same for all of the elements of data in a minibatch.

\subsubsection{Variational dropout}
One limitation of the local reparameterisation trick is that it only applies to networks with no weight sharing, i.e. fully-connected neural networks.
However, inspired by the local reparameterisation trick, a general method for approximating distributions using multiplicative noise can be implemented.
Variational dropout\citep{2015arXiv150602142G,Gal2017ConcreteDropout} is another way of approximating a target distribution, $\rho(\lambda|\chi)$ with $\qa(\lambda|\chi,\w)$.
In this case, the distribution is defined by the application of random variables to the outputs of hidden layers in a neural network.
Much like the local reparameterisation trick, the outputs of the $n_\textrm{l}$ layers of a neural network are draws from some multiplicative noise model, $a^l_{in}=\epsilon_{in}^l\mathcal{f}(o^{l}_{in})$, where $o_{in}^{l}$ are the non-activated outputs of the $l^\textrm{th}$ layer of a neural network at element $n$ in the minibatch. Note that $o_{in}^{l}$ can be obtained using any function, i.e. fully connected, convolutional, etc. $\mathcal{f}$ is some (possibly non-linear) activation function and $\epsilon_{in}^l\sim\textrm{Dist}(m_{in}^l)$ is a random variable drawn from some distribution parameterised by some $\m=\{m_{in}^l|\,i\in[1,\textrm{dim }l],l\in[1,n_\textrm{l}],n\in[1,n_\textrm{elems}]\}$.
Selecting some form for the distribution and values for its parameters, $\m$, provides a way of obtaining samples from the neural network by running the network forward with many draws of $\epsilon$.
This makes the network a model of a Bayesian neural network rather than a Bayesian neural network itself - there is no sampling of the parameters of the network, and no attempt to characterise their uncertainty.
Furthermore, the value of $\m$ cannot be fit using Bayes by Backprop and it is an \emph{a priori} choice for the sampling distribution\footnote{It should be noted that the parameters, $\m$, of any variational dropout distribution can be optimised via expectation maximisation.}.

\paragraph{Bernoulli dropout} One method of performing variational dropout is by using $\boldsymbol{\epsilon}\sim\textrm{Bernoulli}(\m)$, which amounts to feeding forward an input to a network with dropout\cite{Hinton2012} with a keep rate $\m$ for each of the outputs of each layer of the neural network multiple times.
The outputted samples can then be interpreted as the distribution of possible targets which can be obtained using that network (and the choice of the Bernoulli distribution with parameters $\m$).
It is very common to set all values of $m_{in}^l\in\m$ to the same value, although it can be optimised via expectation maximisation.
The ease with which this method can be implemented has made it very popular, and in the limit of large number of samples, the activated outputs approach a Gaussian distribution thanks to the central limit theorem.
Note that the choice of a Bernoulli distribution changes the expected output of any activation layer as $\langle a^l_{in}\rangle=m_{in}^l(1-m_{in}^l)a^l_{in}$, therefore there is a scaling which needs to be taken into account.

\paragraph{Gaussian dropout}
A second option is to draw the random variable from a unit-mean Gaussian\citep{KingmaVariationalTrick, Houthooft2016VIME:Exploration}, $\boldsymbol{\epsilon}\sim\mathbb{N}(\mathbb{I},\textrm{diag}(\m))$, so that the expectation value of the multiplication of the output of a unit of a layer by the random variable remains, $\langle a^l_{in}\rangle=a^l_{in}$ the same since $\langle\epsilon\rangle=1$.
Furthermore, by calculating the variational objective the value of $\m$ in the multiplicative noise distribution can be fit using expectation maximisation.

\medskip

For both the Bernoulli, Gaussian or any other multiplicative dropout distribution, by maximising the $\textrm{ELBO}(\qa)$, we can get $\qa(\lambda|\chi,\w)$ close to $\rho(\lambda|\chi)$ allowing us to make estimates of this distribution.
Again it should be stated that this is not Bayesian in the sense that if the variational distribution provided by variational dropout is approximating the posterior predictive distribution, $\p(\y|\x)$, there is no sense of certainty in how good that approximation is.
There is no attempt to characterise our lack of knowledge of the parameters of the network \emph{or} the parameters of the distributions, $\m$\footnote{Of course, if considering a MAP estimate, then some characterisation of our lack of knowledge is taken into account, but the distribution is still neglected.}.

\subsubsection{Monte Carlo Dropout}

Very closely related to Bernoulli variational dropout, is the MC Dropout model\citep{2015arXiv150602142G}. Completely similar to the previous section, MC Dropout provides a Bayesian framework to interpret the effect of traditional dropout\citep{Hinton2012} on neural networks. A variational distribution $\qa(\w|\chi, \lambda)$ assumed for the network parameter posterior can be parameterised as $\w = \mathbf{M} \cdot  \mathrm{diag}([z_j]_{j=1}^J)$ with $z_j \sim \mathrm{Bernoulli(\m)}$,  $\mathbf{M}$ being a $K \times J$ weight matrix, and $\m$ being the dropout rate. Given this formulation for the variational distribution it can be shown that a KL divergence with respect to an implicit prior can be approximated as a simple $\ell_2$ regularisation term\citep{2015arXiv150602142G}. Training a neural network under dropout and with $\ell_2$ weight regularization therefore maximising the $\textrm{ELBO}(\qa)$ and is performing proper variational inference at no extra cost.

\paragraph{Example: Probabilistic classification of galaxy morphologies and active learning} MC Dropout is the most frequent solution adopted for probabilistic modelling using neural networks, and was the first such application in astrophysics\citep{2017ApJ...850L...7P}, for a strong gravitational lensing parameter estimation problem. To illustrate the method and its applications on a more recent example\citep{2020MNRAS.491.1554W}, we will consider the problem of classifying galaxy types from cutout images. In the context of modern large galaxy surveys, the challenge is to be able to automatically determine galaxy morphological types without (or with minimal) human visual inspection.

\medskip

Such a study is based on the result of a large citizen science effort asking volunteers to answer a series of questions to characterise the type and morphology of a series of galaxy images. 
The task for the neural network is to predict volunteer responses for some galaxy types of particular interest, $k$. 
These answers are modeled using a binomial distribution, $\mathrm{Bin}(\r, N)$, where $\r$ is the probability of a volunteer providing a positive response, and $N$ the number of volunteers asked to answer the question. Based on this model, a probabilistic prediction model can be built from a neural network estimating the parameter $\r$ from a given image $\x$:
\begin{equation}
	\Lambda =  - \log \mathrm{Bin}(k| \x, \w, \a,  N) + \lambda \parallel \w \parallel_2^2  \;.
\end{equation}

\medskip

\begin{figure}
    \centering
    \begin{minipage}[c]{0.5\textwidth}
    \includegraphics[width=\textwidth]{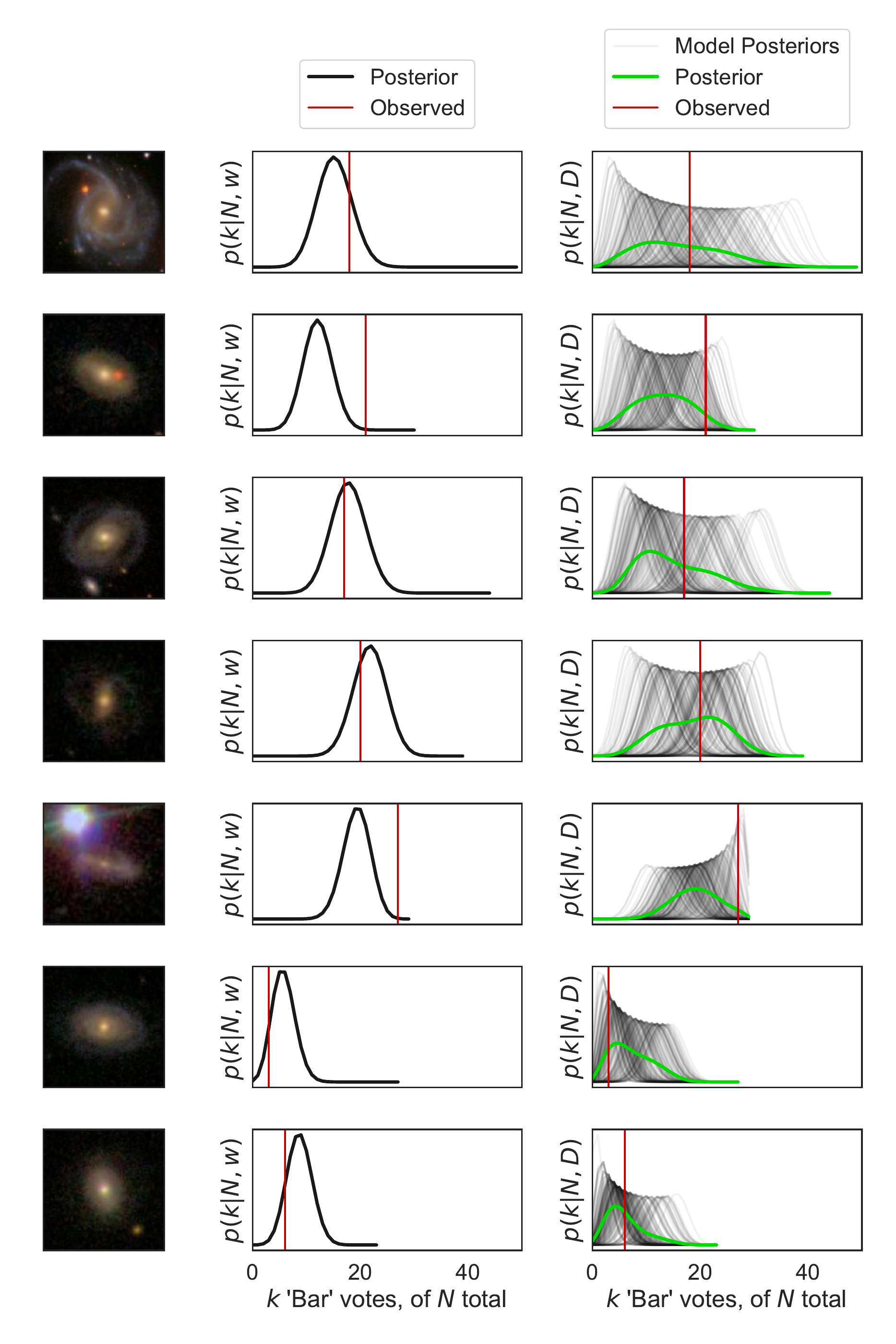}
  \end{minipage}\hfill
  \begin{minipage}[c]{0.5\textwidth}
    \caption{Posterior distributions of number $k$ of positive answers to the question ``Does this image contain a barred spiral galaxy?''  for $N$ votes. Left: Image cutouts as presented to the citizen scientists and CNN. Center: Approximate posterior distribution predicted by one model at fixed network parameters, i.e only modeling aleatoric uncertainties, the red line represents the true observed number. Right: Approximate posterior taking into account a model for both aleatoric and epistemic uncertainties, i.e. obtained by sampling 30 realisations from the MC Dropout network. The full posterior distribution (green) is generally broader and better calibrated than individual approximate posterior samples (black). \textit{Figure credit: Walmsley et al. (2019)\citep{2020MNRAS.491.1554W}}}
    \label{fig:walmsley2019_comparison}
  \end{minipage}
\end{figure}

Figure~\ref{fig:walmsley2019_comparison} illustrates the difference between posterior predictions from the fitted model with and without sampling of network outputs via MC Dropout. For any given fixed realisation (central column), the distribution of predictions is generally over confident, leading to apparent mis-calibration as two out of the seven examples appear to give very low probability to the actual value (second and fifth rows). On the contrary, after sampling from the multiplicative noise distribution (right column), the mean model approximate posterior (green) is significantly wider \emph{and} also can exhibit more complex shapes than a simple binomial distribution. The high variance on the posterior predictions indicates that epistemic uncertainties are more significant, and the authors further measure empirically a much improved, but not perfect, calibration of the MC Dropout posterior.

\medskip

Recognising that the uncertainties modelled by a choice in probabilistic network are not perfect, the
authors still propose an excellent use case for them, in the form of active learning. In the active scenario, approximate and fast inference is preferred over more exact, but computationally and time expensive results.  For this reason, the authors propose a strategy to identify galaxies for which the model uncertainies are largest, and preferentially ask human volunteers to label those, as a way to selectively invest human resources where they will be the most useful to help constrain the model.

\medskip

In particular, the authors adopt the Bayesian Active Learning by Disagreement\citep{Mackay1992} (BALD) strategy which is based on selecting examples that maximize the mutual information $\mathbb{I}[k, \w]$. This quantity measures, for a given galaxy $\x$, how much information can be gained on the parameters of the neural network $\w$ from knowing the true label $k$ of that galaxy. While estimating this mutual information is in general a difficult task, a practical estimator can be derived in the case of a MC dropout. In their experiments, it is found that for some prediction tasks, as much as 60\% fewer training galaxies are necessary to reach a given testing score when selected through active learning, compared to selected through a uniform random sampling.

\subsubsection{Flipout}
Flipout\citep{Wen2018} is an alternative to the local reparameterisation trick (or variational dropout) that proposes an efficient way to generate (and store) pseudo-independent perturbations to decorrelate the gradients with respect to parameters $\m=\{(\mu_i,\Delta\w_i)|i\in[1,n_\textrm{w}]\}$ according to each of the $n_\textrm{elems}$ elements of data within a minibatch. $\mu_i$ and $\Delta\w_i$ are the mean and stochastic perturbation of some network parameter $\w(m_i)$.
This method therefore is more closely akin to stochastic gradient variational Bayes, where the distribution of network parameters is fitted.
Note that for Flipout, the requirements on the distribution for each network parameter is that they are differentiable with respect to the parameters $\m$, that the distribution of perturbations of the network parameters, $\Delta\w_i$, is symmetric about zero (but not necessarily Gaussian) and the network parameters are independent.
With $\Delta\w_i$ symmetric amount zero, the multiplication by a random matrix of signs leaves it identically distributed.
This means that by choosing a single $\Delta\w_i$ for each network parameter (like $\Delta\w_i=\sigma_i\epsilon_i$ for the Gaussian case) somewhat decorrelated gradients can be obtained for each element of a minibatch by identically distributing $\Delta\w_i$ via random draws from two $n_\textrm{elems}$-length vectors, $\mathcal{j}_i=\{j_{in}=2b_{in}-1| b_{in}\sim\textrm{Bernoulli}(0.5), n\in[1,n_\textrm{elems}]\}$ and $\mathcal{k}_i=\{k_{in}=2b_{in}-1| b_{in}\sim\textrm{Bernoulli}(0.5), n\in[1,n_\textrm{elems}]\}$.
Each parameter of the network is then obtained, as with the reparameterisation trick, via
\begin{equation}
	\w_n(m_i)=\mu_{i}+\Delta\w_ij_{in}k_{in}.
\end{equation}
This can be performed very quickly using matrix multiplication, affording a decrease in the variance of the stochastic gradient by a factor of $\sim1/n_\textrm{elems}$ in comparison to using shared parameter values for an entire minibatch for approximately twice the computational cost, although due to parallelisation this can be done in equal time.

\paragraph{Example: Cosmological parameter inference and uncertainty calibration} The inference of cosmological parameter values from data, such as maps of the Cosmic Microwave Background radiation, is an important task in these times of precision cosmology. It is therefore useful to consider the comparison of several of the variational inference methods to calibrate their performance\citep{Hortua2019ParametersNetworks} . The study uses two CNN architectures, AlexNet and VGG to predict, from an image of the CMB, a Gaussian posterior distribution on a limited set of three cosmological parameters. The outputs of these neural networks, parameterised with $\w$, are therefore chosen to be the mean, $\mu\equiv\mu(\x, \w)$, and covariance, $\Sigma\equiv\Sigma(\x, \w)$, of a multivariate Gaussian distribution. The loss function used to train these networks under a Flipout model is
\begin{align}
	\Lambda(\m) =& - \underset{\w\sim\qa}{\mathbb{E}} \left[\frac{1}{2} (\y - \mu)^T \Sigma^{-1}(\y - \mu) + \frac{1}{2} \log\det \Sigma \right] + \mathbb{KL}\left( \qa \parallel \rho \right).
\end{align}

\begin{figure}
\center
\includegraphics[width=0.55\textwidth]{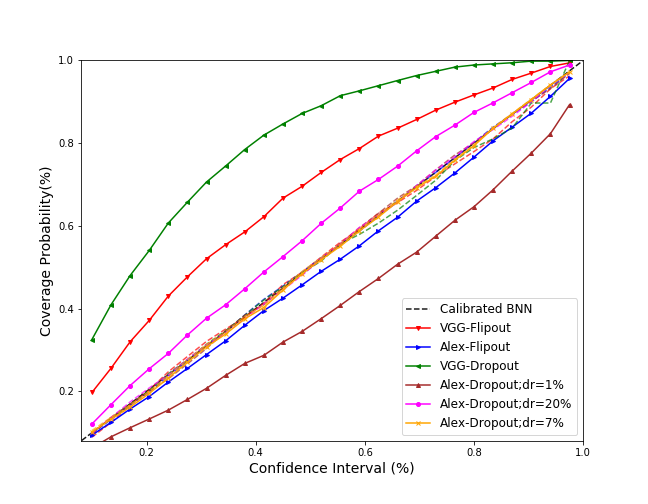}%
\includegraphics[width=0.45\textwidth]{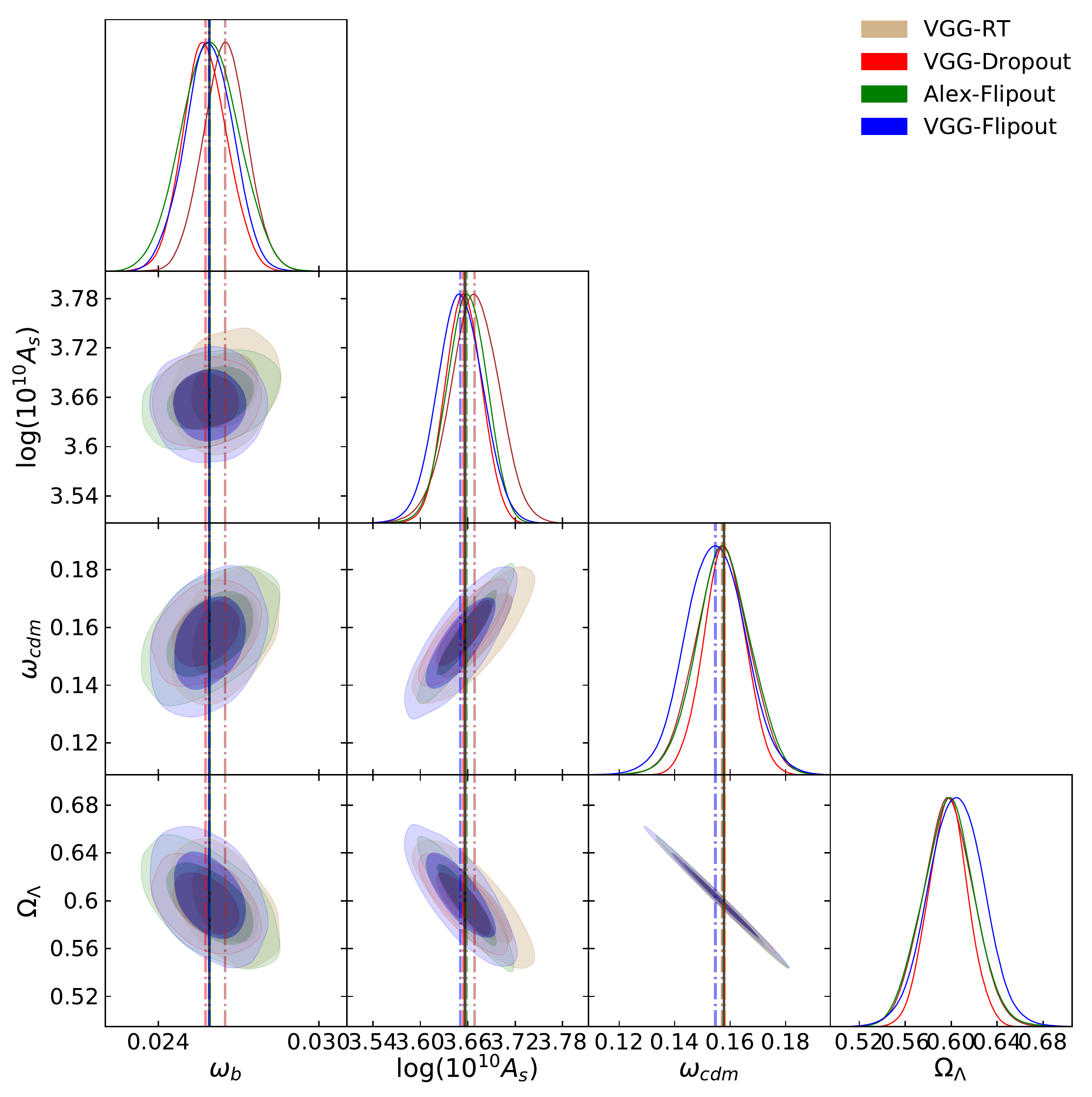}
\caption{Left: Reliability diagrams for different models, before (solid) and after (dashed) post-training re-calibratrion. Right: Approximate posteriors for cosmological parameters obtained after calibration of the models. This illustrates the difficulty of obtaining well calibrated probabilistic models from neural networks directly out of the optimisation procedure, but post-hoc calibration can correct some of these biases. \textit{Figure credit: Hortua et al. (2019)\citep{Hortua2019ParametersNetworks}}}
\label{fig:hortua2019}
\end{figure}

\medskip

In this work, the authors perform a post-training re-calibration of the models to ensure that some coverage properties are respected. In practice, they adopt the Platt Scaling method\citep{Kull2017}, to empirically adjust the posteriors as to make the reliability diagram of their coverage probability well calibrated. Note however that this simple scaling cannot account for all deviations from the true posterior shape.

\medskip

The results of this procedure are illustrated on Figure~\ref{fig:hortua2019} where the left plot shows the reliability diagrams of the various models before and after calibration. The right plot illustrates the confidence contours for the approximate posterior of cosmological parameters predicted by four different re-calibrated models on the same input data. They are fairly similar in terms of sizes, but not identical, showing that this re-calibration cannot account for complex departures in posterior shapes.

\medskip

One of the takeaways of this work is that overall Flipout appear to be the best performing method in terms of calibration, training speed, and accuracy, out of the four explored (reparameterisation, Flipout, MC Dropout, DropConnect).

\subsubsection{Neutra}
We can also use variational inference as part of a Markov chain sampling scheme.
Neutra\cite{Hoffman2019} is a method which samples from a normal distribution and then performs a bijective transformation, $\mathcal{g}:\epsilon\sim\mathbb{N}(\bf{0},\mathbb{I})\to\lambda\sim\rho_\chi$,  to a space approximating the target distribution.
In this way it can be seen as an approximation to the Riemannian manifold HMC (section~\ref{sec:RMHMC}) where the metric is defined by the bijective function, $\mathbb{I}(\lambda)=(\mathcal{J}\mathcal{J}^T)^{-1}$, where the Jacobian is $\mathcal{J}=\partial_\epsilon\mathcal{g}$.
Using neural networks (and particularly many of the modern density estimators such as inverse autoregressive flows, etc.) this Jacobian is very easy to evaluate and therefore $\mathcal{g}$ can be arbitrarily complex, fittable to the desired function by maximising the ELBO and quick to evaluate.
Using HMC, samples can be obtained very easily from the normal distribution and the bijected forward to get samples from an approximation to the target distribution much more efficiently than samples can be obtained by directly evaluating the target distribution.
It should be noted that this method, like all those mentioned in this section, is not Bayesian in nature, since, in this case, there is no quantification in how well the bijection is really performing.
Therefore, there is no way to tell if the samples, $\mathcal{g}(\epsilon)$, actually coincide with samples $\lambda\sim\rho_\chi$, meaning the distribution could be very different from that desired when addressing target distribution using exact evaluations.

\section{Concluding Remarks and Outlook}

Despite impressive accuracy in supervised learning benchmarks, current state of the art neural networks are poor at quantifying predictive uncertainty, and as such are prone to produce overconfident predictions and biases which are extremely difficult to disentangle from true properties of the data. The fact that proper uncertainty quantification is crucial for many practical applications justifies the formulation of neural networks as statistical models as a first step towards using them for inference. 

\medskip 

While truly Bayesian neural networks have the capacity to fully characterize the epistemic uncertainty introduced by the neural network, in practice, exact Bayesian inference is intractable for neural network. 
It is common to resort to either using numerically approximated by exact samples of posterior distribution of network parameters, that is, Monte Carlo methods, or to using approximate distributions as a proxy for the true Bayesian posterior, through variational inference. 
The fact that, through the former method, Bayesian neural networks are often harder to train and implement than non-Bayesian neural networks means that, in the literature, variational methods have gained a lot of popularity in the recent years. 
However, as we have have stressed in this chapter, those latter approximate methods suffer from many pitfalls, in particular the lack of guarantee that the approximate distribution is sufficiently close to the desired target.  

\medskip

Because of these, other statistical tools and tests should be used in concurrence with approximate Bayesian neural networks, such as calibration and test of generalization of the predictive uncertainty to domain shifts\citep{NIPS2019_9547}. However, it is worth noting that Bayesian neural networks are not necessarily the most useful for doing the best reasoned inference of network outputs. For this, other methods, such as likelihood-free (simulation-based) inference, could be more efficient, powerful, and easier to implement.





\bibliographystyle{ws-rv-van}
\bibliography{references}

\begin{thebibliography}{38}
\providecommand{\natexlab}[1]{#1}
\providecommand{\url}[1]{\texttt{#1}}
\expandafter\ifx\csname urlstyle\endcsname\relax
  \providecommand{\doi}[1]{doi: #1}\else
  \providecommand{\doi}{doi: \begingroup \urlstyle{rm}\Url}\fi

\bibitem{athreya2006measure}
K.~Athreya and S.~Lahiri, \emph{Measure Theory and Probability Theory}.
  Springer Texts in Statistics, Springer  (2006).
\newblock ISBN 9780387329031.

\bibitem{Muller1998}
P.~M\"uller and D.~Rios, Issues in bayesian analysis of neural network models,
  \emph{Neural computation}. {\bf 10}, \penalty0 749--70  (03, 1998).

\bibitem{Charnock2019NeuralFunction}
T.~{Charnock}, G.~{Lavaux}, B.~D. {Wandelt}, S.~{Sarma Boruah}, J.~{Jasche},
  and M.~J. {Hudson}, {Neural physical engines for inferring the halo mass
  distribution function}, \emph{Monthly Notices of the Royal Astronomical
  Society}. {\bf 494}\penalty0 (1), \penalty0 50--61  (Mar., 2020).
\newblock \doi{10.1093/mnras/staa682}.

\bibitem{Hafner2018}
D.~Hafner, D.~Tran, T.~Lillicrap, A.~Irpan, and J.~Davidson, {Noise Contrastive
  Priors for Functional Uncertainty}  (2018).
\newblock URL \url{http://arxiv.org/abs/1807.09289}.

\bibitem{NIPS2017_7219}
B.~Lakshminarayanan, A.~Pritzel, and C.~Blundell.
\newblock Simple and scalable predictive uncertainty estimation using deep
  ensembles.
\newblock In eds. I.~Guyon, U.~V. Luxburg, S.~Bengio, H.~Wallach, R.~Fergus,
  S.~Vishwanathan, and R.~Garnett, \emph{Advances in Neural Information
  Processing Systems 30}, pp. 6402--6413. Curran Associates, Inc.  (2017).

\bibitem{PhysRevE.85.026703}
E.~A. J.~F. Peters and G.~de~With, Rejection-free monte carlo sampling for
  general potentials, \emph{Phys. Rev. E}. {\bf 85}, \penalty0 026703  (Feb,
  2012).
\newblock \doi{10.1103/PhysRevE.85.026703}.
\newblock URL \url{https://link.aps.org/doi/10.1103/PhysRevE.85.026703}.

\bibitem{Bouchard2015}
A.~Bouchard-C\^{o}t\'{e}, S.~Vollmer, and A.~Doucet, The bouncy particle
  sampler: a non-reversible rejection-free {M}arkov chain {M}onte {C}arlo
  method  (2015).
\newblock Technical report arxiv:1510.02451.

\bibitem{alex2010optimal}
A.~Beskos, N.~S. Pillai, G.~O. Roberts, J.~M. Sanz-Serna, and A.~M. Stuart.
\newblock Optimal tuning of the hybrid monte-carlo algorithm  (2010).

\bibitem{PhysRevD.100.104023}
Y.~Bouffanais and E.~K. Porter, Bayesian inference for binary neutron star
  inspirals using a hamiltonian monte carlo algorithm, \emph{Phys. Rev. D}.
  {\bf 100}, \penalty0 104023  (Nov, 2019).
\newblock \doi{10.1103/PhysRevD.100.104023}.
\newblock URL \url{https://link.aps.org/doi/10.1103/PhysRevD.100.104023}.

\bibitem{Hoffman2014}
M.~D. Hoffman and A.~Gelman, {The No-U-Turn Sampler : Adaptively Setting Path
  Lengths in Hamiltonian Monte Carlo}. {\bf 15}, \penalty0 1351--1381  (2014).

\bibitem{Lu2017}
X.~Lu, V.~Perrone, L.~Hasenclever, Y.~W. Teh, and S.~J. Vollmer, {Relativistic
  Monte Carlo}, \emph{Proceedings of the 20th International Conference on
  Artificial Intelligence and Statistics, AISTATS 2017}. \penalty0 (1),
  \penalty0 1--11  (2017).

\bibitem{Fu2016}
T.~Fu and Z.~Zhang.
\newblock {Quasi-Newton Hamiltonian Monte Carlo}.
\newblock In \emph{Conference on Uncertainty in Artificial Intelligence (UAI)}
  (2016).

\bibitem{Jasche:2013}
J.~{Jasche} and B.~D. {Wandelt}, {Bayesian physical reconstruction of initial
  conditions from large-scale structure surveys}, \emph{Monthly Notices of the
  Royal Astronomical Society}. {\bf 432}\penalty0 (2), \penalty0 894--913
  (Jun, 2013).
\newblock \doi{10.1093/mnras/stt449}.

\bibitem{Jasche:2015}
J.~{Jasche}, F.~{Leclercq}, and B.~D. {Wandelt}, {Past and present cosmic
  structure in the SDSS DR7 main sample}, \emph{Journal of Cosmology and
  Astroparticle Physics}. 2015\penalty0 (1):\penalty0 036  (Jan, 2015).
\newblock \doi{10.1088/1475-7516/2015/01/036}.

\bibitem{Lavaux:2016}
G.~{Lavaux} and J.~{Jasche}, {Unmasking the masked Universe: the 2M++ catalogue
  through Bayesian eyes}, \emph{Monthly Notices of the Royal Astronomical
  Society}. {\bf 455}\penalty0 (3), \penalty0 3169--3179  (Jan, 2016).
\newblock \doi{10.1093/mnras/stv2499}.

\bibitem{2019PhRvD.100d3515K}
D.~{Kodi Ramanah}, T.~{Charnock}, and G.~{Lavaux}, {Painting halos from cosmic
  density fields of dark matter with physically motivated neural networks},
  \emph{Physical Review D.} 100\penalty0 (4):\penalty0 043515  (Aug., 2019).
\newblock \doi{10.1103/PhysRevD.100.043515}.

\bibitem{Girolami2009}
M.~Girolami, B.~Calderhead, and S.~A. Chin, {Riemannian Manifold Hamiltonian
  Monte Carlo}  (2009).
\newblock URL \url{http://arxiv.org/abs/0907.1100}.

\bibitem{Chen2014}
T.~Chen, E.~B. Fox, and C.~Guestrin.
\newblock {Stochastic gradient Hamiltonian Monte Carlo}.
\newblock In \emph{31st International Conference on Machine Learning, ICML
  2014}, vol.~5, pp. 3663--3676  (feb, 2014).
\newblock ISBN 9781634393973.
\newblock URL \url{http://arxiv.org/abs/1402.4102}.

\bibitem{2015arXiv150505424B}
C.~Blundell, J.~Cornebise, K.~Kavukcuoglu, and D.~Wierstra, {Weight Uncertainty
  in Neural Networks}, \emph{arXiv e-prints}. p. arXiv:1505.05424  (5, 2015).
\newblock URL \url{http://arxiv.org/abs/1505.05424}.

\bibitem{Dikov2019BayesianArchitectures}
G.~Dikov, P.~van~der Smagt, and J.~Bayer, {Bayesian Learning of Neural Network
  Architectures}  (1, 2019).
\newblock URL \url{http://arxiv.org/abs/1901.04436}.

\bibitem{2013arXiv1306.0733K}
D.~P. {Kingma}, {Fast Gradient-Based Inference with Continuous Latent Variable
  Models in Auxiliary Form}, \emph{arXiv e-prints}. art. arXiv:1306.0733
  (June, 2013).

\bibitem{2013arXiv1312.6114K}
D.~P. {Kingma} and M.~{Welling}, {Auto-Encoding Variational Bayes}, \emph{arXiv
  e-prints}. art. arXiv:1312.6114  (Dec., 2013).

\bibitem{2020MNRAS.491.4277M}
A.~M{\"{o}}ller and T.~de~Boissi{\`{e}}re, {SuperNNova: an open-source
  framework for Bayesian, neural network-based supernova classification},
  \emph{Monthly Notices of the Royal Astronomical Society}. {\bf 491}\penalty0
  (3), \penalty0 4277--4293  (1, 2020).
\newblock ISSN 0035-8711.
\newblock \doi{10.1093/mnras/stz3312}.
\newblock URL \url{https://academic.oup.com/mnras/article/491/3/4277/5651173}.

\bibitem{Fortunato2017}
M.~Fortunato, C.~Blundell, and O.~Vinyals, {Bayesian Recurrent Neural
  Networks}. pp. 1--14  (apr, 2017).
\newblock URL \url{http://arxiv.org/abs/1704.02798}.

\bibitem{degroot1983}
M.~H. DeGroot and S.~E. Fienberg, {The Comparison and Evaluation of
  Forecasters}, \emph{Journal of the Royal Statistical Society. Series D (The
  Statistician)}. {\bf 32}\penalty0 (1/2), \penalty0 12--22  (1983).
\newblock ISSN 00390526, 14679884.
\newblock URL \url{http://www.jstor.org/stable/2987588}.

\bibitem{KingmaVariationalTrick}
D.~P. Kingma, T.~Salimans, and M.~Welling.
\newblock {Variational Dropout and the Local Reparameterization Trick}.
\newblock Technical report .

\bibitem{2015arXiv150602142G}
Y.~Gal and Z.~Ghahramani, {Dropout as a Bayesian Approximation: Representing
  Model Uncertainty in Deep Learning}, \emph{arXiv e-prints}. p.
  arXiv:1506.02142  (6, 2015).
\newblock URL \url{http://arxiv.org/abs/1506.02142}.

\bibitem{Gal2017ConcreteDropout}
Y.~Gal, J.~Hron, and A.~Kendall, {Concrete Dropout}, \emph{arXiv e-prints}. p.
  arXiv:1705.07832  (5, 2017).

\bibitem{Hinton2012}
G.~E. Hinton, N.~Srivastava, A.~Krizhevsky, I.~Sutskever, and R.~Salakhutdinov,
  Improving neural networks by preventing co-adaptation of feature detectors,
  \emph{CoRR}. {\bf abs/1207.0580}  (2012).
\newblock URL \url{http://arxiv.org/abs/1207.0580}.

\bibitem{Houthooft2016VIME:Exploration}
R.~Houthooft, X.~Chen, Y.~Duan, J.~Schulman, F.~De~Turck, and P.~Abbeel, {VIME:
  Variational Information Maximizing Exploration}  (5, 2016).
\newblock URL \url{http://arxiv.org/abs/1605.09674}.

\bibitem{2017ApJ...850L...7P}
L.~Perreault~Levasseur, Y.~D. Hezaveh, and R.~H. Wechsler, {Uncertainties in
  Parameters Estimated with Neural Networks: Application to Strong
  Gravitational Lensing}, \emph{The Astrophysical Journal}. {\bf 850}\penalty0
  (1), \penalty0 L7  (11, 2017).
\newblock ISSN 2041-8213.
\newblock \doi{10.3847/2041-8213/aa9704}.
\newblock URL
  \url{http://stacks.iop.org/2041-8205/850/i=1/a=L7?key=crossref.dd8f01b687a77b74ce33239cdb39c453}.

\bibitem{2020MNRAS.491.1554W}
M.~Walmsley, L.~Smith, C.~Lintott, Y.~Gal, S.~Bamford, H.~Dickinson,
  L.~Fortson, S.~Kruk, K.~Masters, C.~Scarlata, B.~Simmons, R.~Smethurst, and
  D.~Wright, {Galaxy Zoo: probabilistic morphology through Bayesian CNNs and
  active learning}, \emph{Monthly Notices of the Royal Astronomical Society}.
  {\bf 491}\penalty0 (2), \penalty0 1554--1574  (1, 2020).
\newblock ISSN 0035-8711.
\newblock \doi{10.1093/mnras/stz2816}.
\newblock URL \url{https://academic.oup.com/mnras/article/491/2/1554/5583078}.

\bibitem{Mackay1992}
D.~J.~C. MacKay, {Information-Based Objective Functions for Active Data
  Selection}, \emph{Neural Computation}. {\bf 4}\penalty0 (4), \penalty0
  590--604  (1992).
\newblock ISSN 0899-7667.
\newblock \doi{10.1162/neco.1992.4.4.590}.

\bibitem{Wen2018}
Y.~Wen, P.~Vicol, J.~Ba, D.~Tran, and R.~Grosse, {Flipout: Efficient
  pseudo-independent weight perturbations on mini-batches}, \emph{6th
  International Conference on Learning Representations, ICLR 2018 - Conference
  Track Proceedings}. pp. 1--16  (2018).

\bibitem{Hortua2019ParametersNetworks}
H.~J. Hortua, R.~Volpi, D.~Marinelli, and L.~Malag{\`{o}}, {Parameters
  Estimation for the Cosmic Microwave Background with Bayesian Neural
  Networks}, \emph{arXiv e-prints}. p. arXiv:1911.08508  (11, 2019).

\bibitem{Kull2017}
M.~Kull, T.~d.~M. {e Silva Filho}, and P.~Flach, {Beta calibration: A
  well-founded and easily implemented improvement on logistic calibration for
  binary classifiers}, \emph{Proceedings of the 20th International Conference
  on Artificial Intelligence and Statistics, AISTATS 2017}. {\bf 54}  (2017).

\bibitem{Hoffman2019}
M.~{Hoffman}, P.~{Sountsov}, J.~V. {Dillon}, I.~{Langmore}, D.~{Tran}, and
  S.~{Vasudevan}, {NeuTra-lizing Bad Geometry in Hamiltonian Monte Carlo Using
  Neural Transport}, \emph{arXiv e-prints}. art. arXiv:1903.03704  (Mar.,
  2019).

\bibitem{NIPS2019_9547}
J.~Snoek, Y.~Ovadia, E.~Fertig, B.~Lakshminarayanan, S.~Nowozin, D.~Sculley,
  J.~Dillon, J.~Ren, and Z.~Nado.
\newblock Can you trust your model's uncertainty? evaluating predictive
  uncertainty under dataset shift.
\newblock In eds. H.~Wallach, H.~Larochelle, A.~Beygelzimer, F.~d'Alch\'{e}
  Buc, E.~Fox, and R.~Garnett, \emph{Advances in Neural Information Processing
  Systems 32}, pp. 13991--14002. Curran Associates, Inc.  (2019).

\end{thebibliography}


\end{document}